\documentclass[preprint,12pt]{elsarticle}

\usepackage{amsmath}
\usepackage{float}
\usepackage{graphicx}
\usepackage{makecell}
\usepackage[dvipsnames,svgnames,x11names]{xcolor}
\usepackage{makecell}
\graphicspath{{./Figures/}}

\journal{Journal of Computational Physics}

\begin{document} 


\title{Exact Enforcement of Temporal Continuity in Sequential Physics-Informed Neural Networks}
\author[add1]{Pratanu Roy\corref{cor1}}
\ead{roy23@llnl.gov}
\cortext[cor1]{Corresponding authors. These authors contributed equally to this work.}

\author[add2]{Stephen Castonguay\corref{cor1}}
\ead{castonguay1@llnl.gov}
\address[add1]{Atmospheric, Earth and Energy Division, Lawrence Livermore National Laboratory, \\ 7000 East Avenue, Livermore, CA 94550, USA}
\address[add2]{Computational Engineering Division, Lawrence Livermore National Laboratory, \\ 7000 East Avenue, Livermore, CA 94550, USA}
\date{}
\begin{frontmatter}
\begin{abstract}
The use of deep learning methods in scientific computing represents a potential paradigm shift in engineering problem solving. 
One of the most prominent developments is Physics-Informed Neural Networks (PINNs), in which neural networks are trained to satisfy partial differential equations (PDEs). While this method shows promise, the standard version has been shown to struggle in accurately predicting the dynamic behavior of time-dependent problems. To address this challenge, methods have been proposed that decompose the time domain into multiple segments, employing a distinct neural network in each segment and directly incorporating continuity between them in the loss function of the minimization problem.
In this work we introduce a method to exactly enforce continuity between successive time segments via a solution ansatz. This hard constrained sequential PINN (HCS-PINN) method is simple to implement and eliminates the need for any loss terms associated with temporal continuity. The method is tested for a number of benchmark problems involving both linear and non-linear PDEs. Examples include various first order time dependent problems in which traditional PINNs struggle, namely advection, Allen-Cahn, and Korteweg-de Vries equations.
Furthermore, second and third order time-dependent problems are demonstrated via wave and Jerky dynamics examples, respectively.
Notably, the Jerky dynamics problem is chaotic, making the problem especially sensitive to temporal accuracy.  
The numerical experiments conducted with the proposed method demonstrated superior convergence and accuracy over both traditional PINNs and the soft-constrained counterparts. 
\end{abstract}


\end{frontmatter}

\section{Introduction}
Physics-Informed Neural Networks, introduced by Raissi et al.~\cite{raissi}, encompass a general framework that can incorporate physical constraints via differential equations. 
In general, it involves using a deep neural network as a trial solution to a PDE-constrained optimization problem whose loss function includes residuals from governing equations, boundary and initial conditions, as well as observed data, if present. 
It is meshless, and the recent progress in deep learning tools has made its implementation through automatic differentiation a straightforward exercise. Indeed, it has been used in modeling a variety of problems, ranging from fluid mechanics \cite{cai2021physics, mahmoudabadbozchelou2022nn,eivazi2022physics,biswas2023three}, heat transfer \cite{cai2021heat}, solid mechanics~\cite{ghaffari2023deep} to subsurface systems \cite{sarma2023variational} and chemical processes \cite{niaki2021physics}. 
Additionally, it shows promise for inverse problems as it can readily accommodate parameters and data into its structure \cite{tanyu2023deep, yang2021b,lu2021physics,jagtap2022deep,chen2020physics,serebrennikova2024physics}.

However, it has not seen widespread adoption in the computational science community. For one, it generally is not as accurate as traditional numerical methods such as the finite element and finite volume methods \cite{reddy2022finite}. 
Additionally, the standard PINN method is not robust as it often fails to converge due to the complexity of the loss landscape \cite{krishnapriyan2021characterizing}. Moreover, time-dependent problems can suffer from a causality violation when the temporal variable is treated the same way as spatial variables \cite{wang2022respecting}. Therefore, various improvements must be made before it can be an acceptable alternative for traditional numerical methods \cite{cuomo2022scientific}.       
One common approach to simplify the process is to directly enforce initial and/or boundary conditions via specific solution ansatz or modified neural network architecture. This leaves only the PDE residuals (and possibly observable data) as terms in the loss function. For time-dependent problems, an additional approach is to divide the temporal domain into segments and use different neural networks for each one. 
This takes advantage of the inherent unidirectional nature of time and helps avoid spurious solutions by maintaining causality.

In this work, we propose a novel method for exactly imposing continuity between successive time segments, leading to hard constrained sequential PINNs (HCS-PINNs).  We demonstrate its effectiveness for a set of carefully chosen time-dependent problems with increasing difficulty. The problem set encompasses both linear and non-linear equations involving first, second and third order derivatives in time. As demonstrated in the results, the HCS-PINN method can accurately predict the solution of time-dependent problems for which traditional PINNs struggle. 

The outline of the paper is as follows. Section \ref{sec:background} presents the numerical background of PINNs, describing various temporal strategies and their advantages and limitations. Section \ref{sec:hcspinn} describes our proposed method as a hard constrained strategy that preserves temporal continuity. In Section \ref{sec:results}, a number of time-dependent benchmark problems are solved using the proposed method and their solution accuracy is compared with the soft constrained counterpart. We conclude the paper by summarizing the findings in Section \ref{sec:summary}.

\section{Background}\label{sec:background}
\begin{figure} [H]
	\centering
	\includegraphics[width=0.8\textwidth]{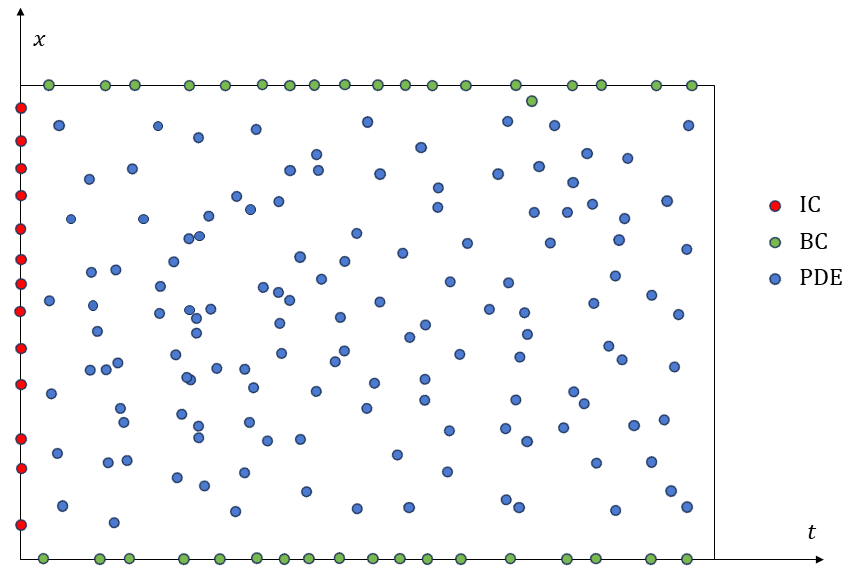}
	\caption{Collocation points for PDE, initial and boundary conditions.}
	\label{fig:Collocation}
\end{figure}

Consider the general representation of a partial differential equation:
\begin{equation}
\mathcal{F}[u(x,t)]= 0,  \quad x\in \Omega,\: t \in [0,T],
\end{equation}
where $\mathcal{F}$ is an operator which may consist of time and spatial derivatives. This is accompanied by
\begin{align}
	\mathcal{I}[u(x,0)] &= 0,  \quad x\in \Omega, \\ 
	\mathcal{B}[u(x,t)] &= 0,  \quad x\in \partial \Omega,\: t \in [0,T],
\end{align}
where $\mathcal{I}$ and $\mathcal{B}$ are initial and boundary condition operators, respectively.
The vanilla version of PINNs involves a minimization problem with both initial and boundary conditions included in the loss function in addition to the PDE residual, i.e.
\begin{equation}
	u(x,t) \approx f(x,t,\theta); \quad \theta = \arg \min_{\theta^{*}} \mathcal{L}(\theta^{*}),
\end{equation}
where $f(x,t,\theta)$ is the output of a neural network with $x$ and $t$ as input neurons and $\theta$ as weights and biases. The loss terms are given by
\begin{align}
	\mathcal{L}({\theta}) &= \lambda_P\mathcal{L}_{\text{P}}({\theta})+\lambda_I\mathcal{L}_{\text{I}}({\theta})+\lambda_B\mathcal{L}_{\text{B}}({\theta}), \\
    \mathcal{L}_{\text{P}}({\theta})&=\frac{1}{N_{P}}\sum_{i=1}^{N_{P}} (\mathcal{F}[f(x_i,t_i,\theta)])^2, \quad x_i \in \Omega,\: t_i \in [0,T], \\
    \mathcal{L}_{\text{I}}({\theta})&=\frac{1}{N_{I}}\sum_{i=1}^{N_{I}} (\mathcal{I}[f(x_i,0,\theta)])^2,  \quad x_i \in \Omega, \\ 
	\mathcal{L}_{\text{B}}({\theta})&=\frac{1}{N_{B}}\sum_{i=1}^{N_{B}} (\mathcal{B}[f(x_i,t_i,\theta)])^2, \quad x_i \in \partial \Omega,\: t_i \in [0,T].
\end{align}
$N_P$, $N_I$, and $N_B$ are the number of collocation points for the PDE, initial and boundary conditions, respectively. Additionally, $\lambda_P$, $\lambda_I$, and $\lambda_B$ are parameters which weight the individual impact of each loss term. Figure \ref{fig:Collocation} shows a sample distribution of random collocation points for a two-dimensional spatio-temporal domain.

\subsection{Soft Versus Hard Constraints}
The above formulation enforces the initial and boundary conditions in a soft manner, which leads to a competition between driving down the PDE residual and satisfying the boundary and initial conditions. To obtain satisfactory solutions, the weights must often be adjusted to balance the losses among these three terms. To circumvent this, various strategies have been proposed that exactly enforce these conditions (i.e. in a hard manner), leaving only the PDE residual to be minimized. 

Periodic boundary conditions can be enforced exactly by introducing an intermediate layer between the input neurons and the first hidden layer \cite{dong2021method}. Initial, Dirichlet, Neumann, and Robin boundary conditions can be exactly enforced via solution ansatz that includes the output of the neural network \cite{lu2021physics}. For example, the following simple form exactly satisfies the initial condition for first order problems in time:
\begin{equation}
	u(x,t)= g_0(x) + f(x,t,\theta)t,
\end{equation}
where $u(x,0)=g_0(x)$ is the given initial condition.
For simple geometric domains, analytic ansatz can generally be found to enforce boundary conditions as well. However, the situation becomes much more complicated for arbitrarily complex geometries, although some recent progress has been made in this area. For instance, Sukumar and Srivastava \cite{sukumar2022exact} generated approximate composite distance functions to exactly enforce both Dirichlet and Neumann conditions.
\subsection{Temporal Strategies}
For time-dependent problems, vanilla PINNs often suffer from error propagation due to the unidirectional nature of time, leading to nonphysical results. Because of inherent causality, time should not be treated similarly as other space variables. Various methods have been proposed to address this issue. 

One class of methods involve decomposing the domain into multiple time segments while using a single global neural network as the trial solution.
For instance, Wight and Zhao \cite{wight2020solving} introduced the Adaptive Time Sampling approach. The neural network is first trained on a small initial subdomain. After convergence, the training domain is extended by a time increment, and the network is retrained for the new extended time interval. This continues until the neural network is trained over the whole domain. A similar method was proposed by Mattey and Ghosh \cite{mattey2022novel}, in which each time slab is trained sequentially, with an additional term added to the loss function so that the solution remains compatible with all previous times steps. This method is known as backward compatible PINNs (bc-PINNs). Wang et al. \cite{wang2022respecting} developed the Causal Training approach, in which time steps are used to weight each subdomain separately. The weights are defined in terms of the loss from all prior time steps, ensuring that for any time step the solution up to that point is obtained with adequate accuracy.

Time stepping schemes have also been used in which a separate neural network is defined for each time slab \cite{wight2020solving,krishnapriyan2021characterizing,bihlo2022physics,wang2023expert}. In this case, the initial condition for each segment depends on the PINN solution at the end of the previous time step. Penwarden et al.~\cite{penwarden2023unified} unified many of these ideas under the XPINNs framework, which allows multiple NNs to be trained either sequentially, all at once, or through a moving window strategy in which multiple are trained at the same time. 

Each of the previous methods that utilize multiple neural networks enforce temporal compatibility conditions via loss terms in the minimization problem, leading to only approximate continuity between time steps. In this sense, they can be categorized as Soft Constrained Sequential PINNs (SCS-PINNs). As shown in the next section, temporal continuity can be imposed exactly via a solution ansatz leading to Hard Constrained Sequential PINNs.

\section{Hard Constrained Sequential PINNs (HCS-PINNs)}\label{sec:hcspinn}
\begin{figure} [h]
	\centering
	\includegraphics[width=0.9\textwidth]{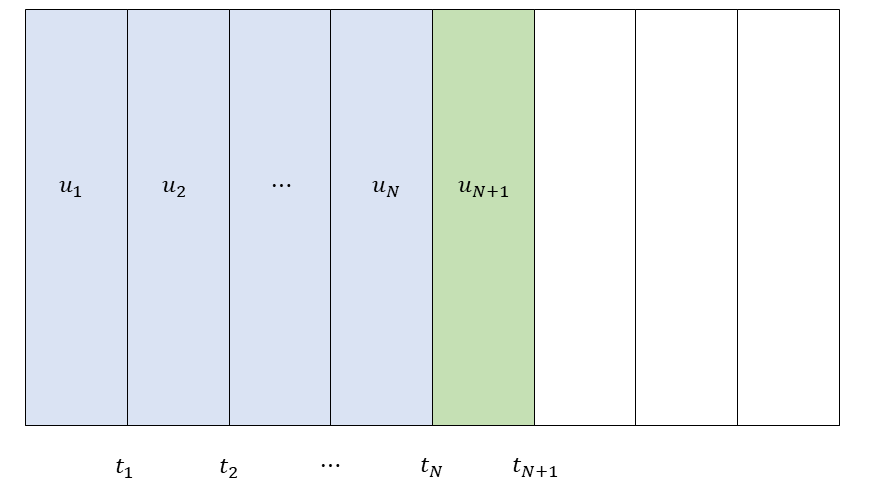}
	\caption{Neural network solution in multiple time sub-domains.}
	\label{fig:bddt}
\end{figure}

\begin{figure} [h]
	\centering
	\includegraphics[width=0.495\textwidth]{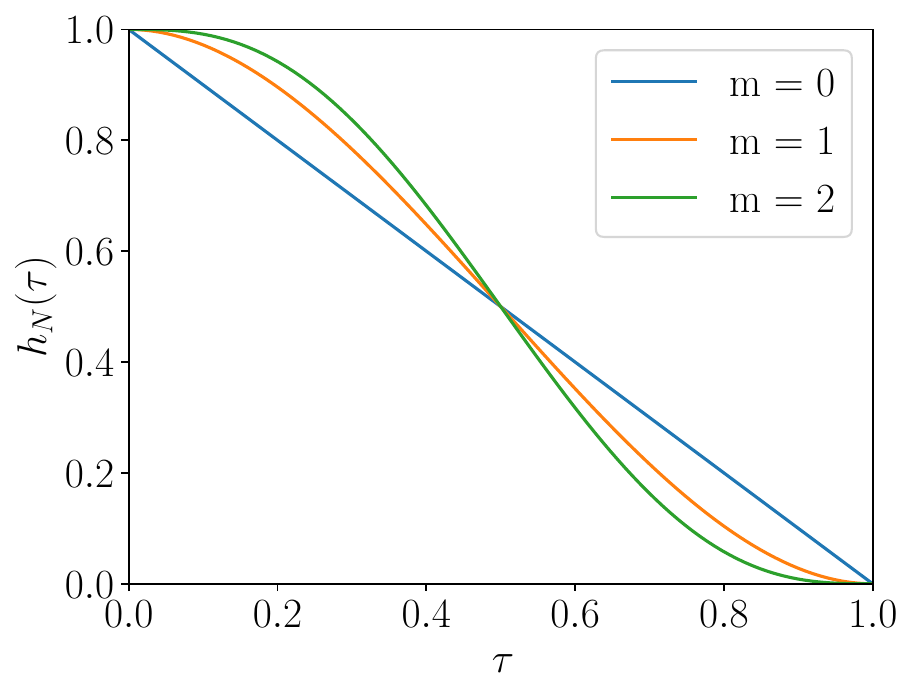}
 	\includegraphics[width=0.495\textwidth]{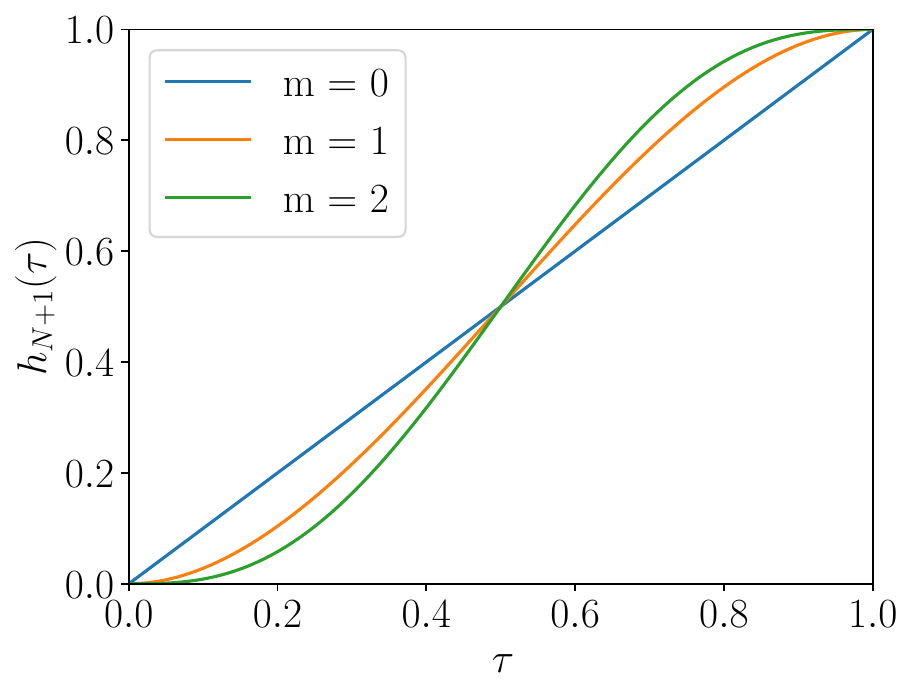}
	\caption{Interpolation functions for $C^0$, $C^1$, and $C^2$ continuity.}
	\label{fig:interp_func}
\end{figure}

First, the temporal domain is decomposed into time windows, as shown in Figure \ref{fig:bddt}. The beginning and end points of a given time window are denoted as $t_{N}$ and $t_{N+1}$, respectively. The solution in this sub-domain is denoted as $u_{N+1}$.
A straightforward approach to enforcing $C^0$ continuity would be to use
\begin{align}
	u_{1}(x,t,\theta_{1}) &= g(x,t) + f(x,t,\theta_1)(t), \\
	u_{N+1}(x,t,\theta_{N+1}) &= u_N(x, t,\theta_{N}) + f(x,t,\theta_{N+1})(t-t_N) \quad \forall N \ge 1 
\end{align}
Where $g(x,t)$ satisfies $\mathcal{I}[g(x,0)]=0 $.
However, this is a recursive form, where neural network solutions for all previous time windows must be evaluated to compute the solution for the current time window.  In order to avoid a recursive definition for the trial solution, we propose the following:  
\begin{align}
    u_{1}(x,t,\theta_{1}) &= h_N(t) g(x,t) + h_{N+1}(t)f_1(x,t,\theta_1), \label{eq:initial}\\
    u_{N+1}(x,t,\theta_{N+1})&= h_N(t)  f_N(x,t,\theta_{N})  \notag\\
        &+h_{N+1}(t) f_{N+1}(x,t,\theta_{N+1})\quad \forall N \ge 1, \label{eq:N+1}
\end{align}
Here, $h_N$ and $h_{N+1}$ are auxiliary functions chosen to obey certain conditions depending on the desired continuity of time derivatives. In order to break recursion and maintain $C^0$ continuity, we define  $\tau= \frac{t-t_N}{t_{N+1}-t_N} $ and require the following:
\begin{align}
    h_N(\tau=0) &= 1,\: h_N(\tau=1) = 0, \\
    h_{N+1}(\tau=0) &= 0,\: h_{N+1}(\tau=1) = 1. 
\end{align}
$C^{m}$ continuity can be obtained by additionally enforcing that the derivatives up to order $m$ vanish at $\tau =0$ and $\tau=1$. The auxiliary functions can be chosen by including polynomials from order 0 up to order $2m+1$ while enforcing these boundary conditions.
Generally speaking, the continuity requirement between time segments is one less than the highest order time derivative in the problem, and choosing a value greater than this will unnecessarily over-constrain the problem.
In this work, we will consider first, second, and third order problems with respect to time, which therefore leads to the choices for $h(\tau)$ listed in Table \ref{IntFunctions}. Note that these functions satisfy a partition of unity, and can be interpreted as interpolation functions (see Figure \ref{fig:interp_func}). 
\begin{table}[h] 
\caption{Auxiliary or Interpolation Functions}
\begin{center}
\begin{tabular}{ |c|c|c| } 
 \hline
 $C^m$  & $h_N(\tau)$ & $h_{N+1}(\tau)$\\  \hline
  $m = 0$ & $1-\tau$ & $\tau$ \\
  $m = 1$ & $1-3\tau^2+2\tau^3$ & $3\tau^2-2\tau^3$ \\
  $m = 2$ & $1-6\tau^5+15\tau^4-10\tau^3$ & $6\tau^5-15\tau^4+10\tau^3$ \\
 \hline
\end{tabular}
\end{center}\label{IntFunctions}
\end{table}

While these functions enforce continuity between time steps, the initial condition must be satisfied as well. Consider initial conditions given in the form: 
\begin{equation}
	\begin{split}
u(x,0) &= g_0(x) \\
u_t(x,0) &= g_1(x) \\
u_{tt}(x,0) &= g_2(x) \quad \\
& \vdots
	\end{split}
\end{equation}
The initial conditions for $M^{\text{th}}$-order problems in time (requiring $C^{M-1}$ continuity) are satisfied by taking the first $M$ terms of the following series:
\begin{equation}
g(x,t) = g_0(x) + tg_1(x) +\frac{t^2}{2}g_2(x)+ \ldots  \frac{t^n}{n!}g_n(x)  +\, \ldots
\end{equation}

\section{Results}\label{sec:results}
A framework was developed to test the accuracy and robustness of physics informed neural network algorithms for dynamic problems. The Python based Jax library \cite{jax2018github} was used to develop the code, which employs feed forward neural networks. For all problems, boundary conditions are enforced directly in the solution ansatz or encoded in the neural network architecture (see \ref{Appendix:PBC}). Therefore, the loss terms for HCS-PINNs solely consist of PDE residuals, whereas for SCS-PINNs, they include both initial and temporal continuity conditions. A dense distribution of random collocation points is used to sample each time segment, with mini-batching employed to speed up the training time, except for the Jerk equation where a full batch gradient descent is used. The optimization procedure starts with a fixed number of Adam \cite{kingma2014adam} iterations. Then the L-BFGS \cite{liu1989limited} iterations are performed until a convergence criterion is met or until it reaches a maximum number of iterations. Due to the stochastic nature of the mini-batching procedure, a simple five term moving average of the change in loss function from iteration to iteration is used as the convergence criterion. Additionally, a different (larger) mini-batch is used to evaluate the loss criterion to prevent false convergence. The activation function is the hyperbolic tangent ($tanh$), with all parameters initialized using the Glorot scheme \cite{glorot2010understanding}. An empirical linear loss weighting scheme is used to further enforce temporal causality (see \ref{EmpWeight}). For any problem without an analytical solution, reference solutions were generated with the Chebfun package \cite{driscoll2014chebfun} in Matlab. In each scenario, we analyze the relative $L^2$ error and training duration between HCS-PINN and SCS-PINN. The simulations were conducted using an Apple M1 Max processor, featuring a clock speed of 3.2 GHz and a memory of 32 GB. 

We conduct numerical experiments by solving five dynamic problems with increasing level of complexity. For each problem, three scenarios are presented in which the temporal decomposition varies by the number of time steps ($nt$). The first two problems, the advection equation and wave equation, are linear and exhibit hyperbolic behavior. Despite their linear nature, the presence of a high characteristic velocity poses challenges for predicting solution over extended time intervals. The subsequent two problems, namely Allen-Cahn and  Korteweg-de Vries (KdV) equations, are non-linear and commonly used in the PINN literature to benchmark algorithms. Lastly, we address a  Jerky dynamics equation, known for its chaotic behavior.  

\subsection{Advection Equation}
The advection equation describes transport of a quantity that moves with a fluid. The periodic, one-dimensional version that is used in many PINN papers is given by 
\begin{align} \label{AdvecEqn}
	u_{t} + c u_{x} &= 0 , x \in [0, 2\pi], t \in [0,1] \\
	u(0,x) &= sin(x) \\
	u(t,0) &= u(t,2\pi)
\end{align}
where $c$ is the prescribed fluid velocity. The exact solution to this problem is given by
\begin{equation}
u(x,t) = sin(x-ct)
\end{equation}
As the advection equation is first order in time, $C^0$ continuity needs to be enforced between adjacent time segments. In our implementation, the PDE has been normalized by the fluid velocity. We found that this normalization sufficiently enforced the initial condition for SCS-PINNs. Alternatively, the penalty parameter $\lambda_I$ could be increased to achieve the same effect. 

\begin{figure} 
\centering
\begin{tabular}{c c}
  HCS-PINN, c = 30 &SCS-PINN, c = 30  \\ 
\includegraphics[trim={0.25cm 0 0.25cm 0},clip, width=0.46\linewidth]{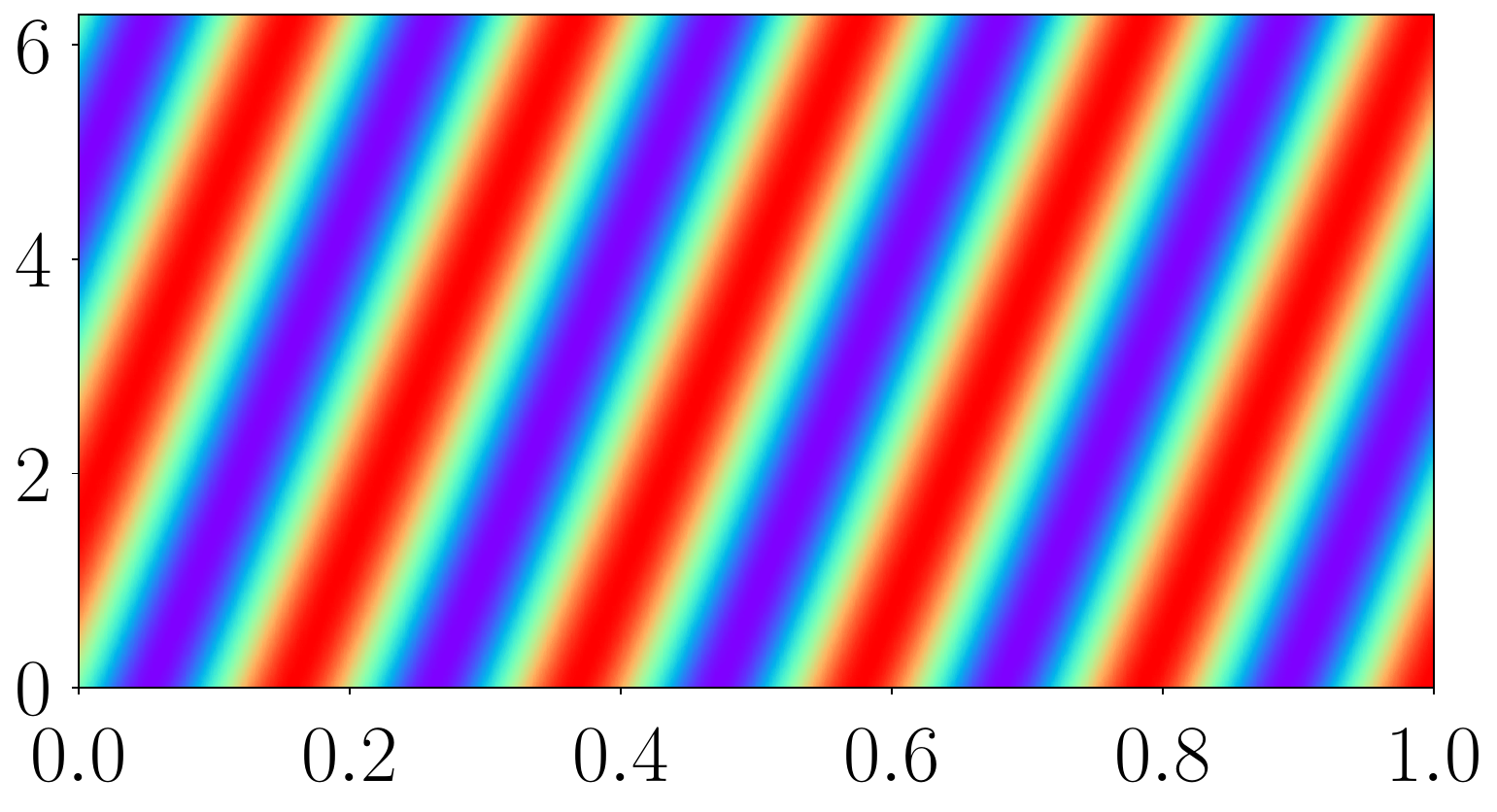}&
\includegraphics[trim={0.25cm 0 0.25cm 0},clip, width=0.46\linewidth]{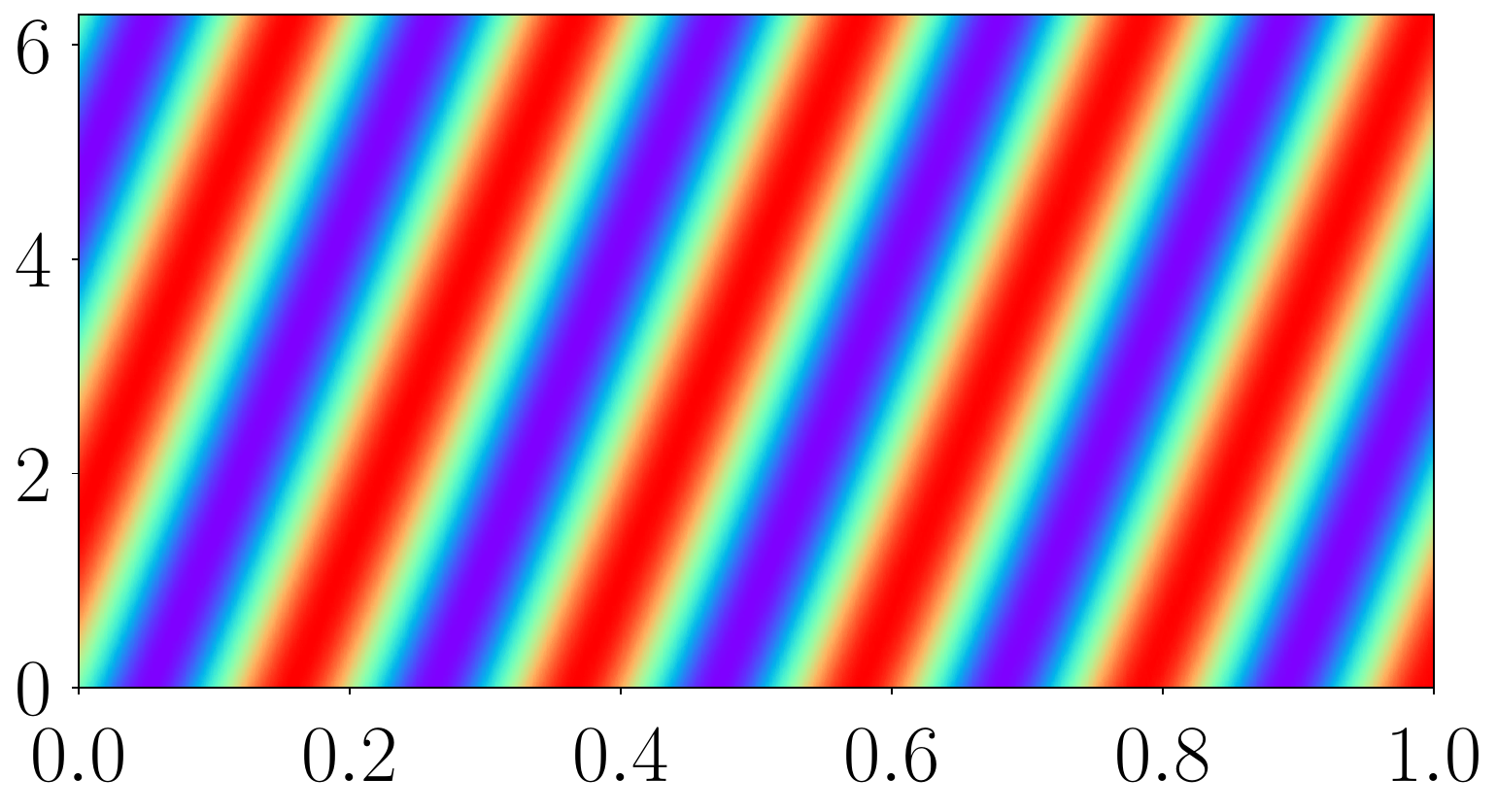}\\
 HCS-PINN, c = 50 &SCS-PINN, c = 50  \\
\includegraphics[trim={0.25cm 0 0.25cm 0},clip, width=0.46\linewidth]{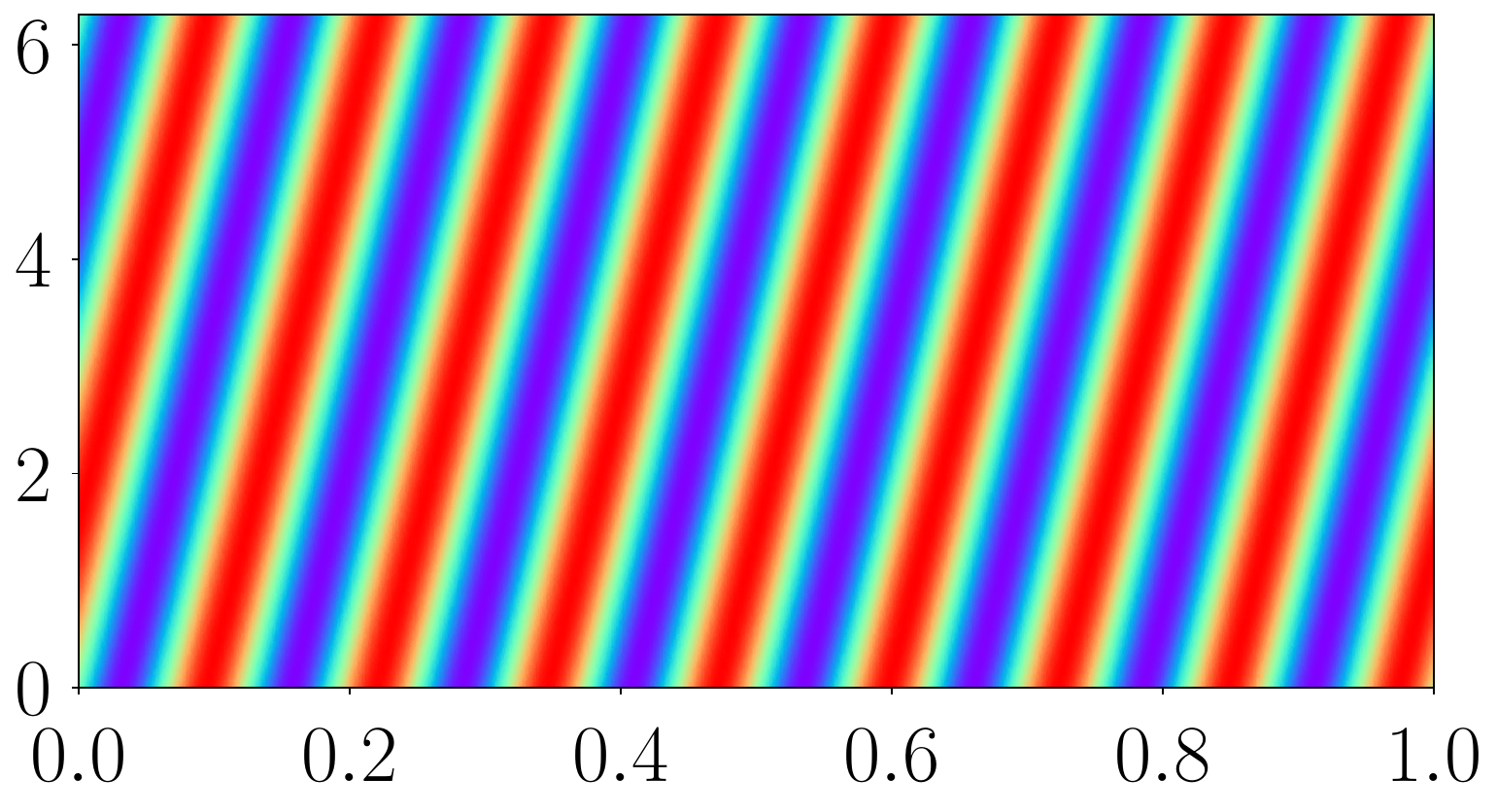}&
\includegraphics[trim={0.25cm 0 0.25cm 0},clip, width=0.46\linewidth]{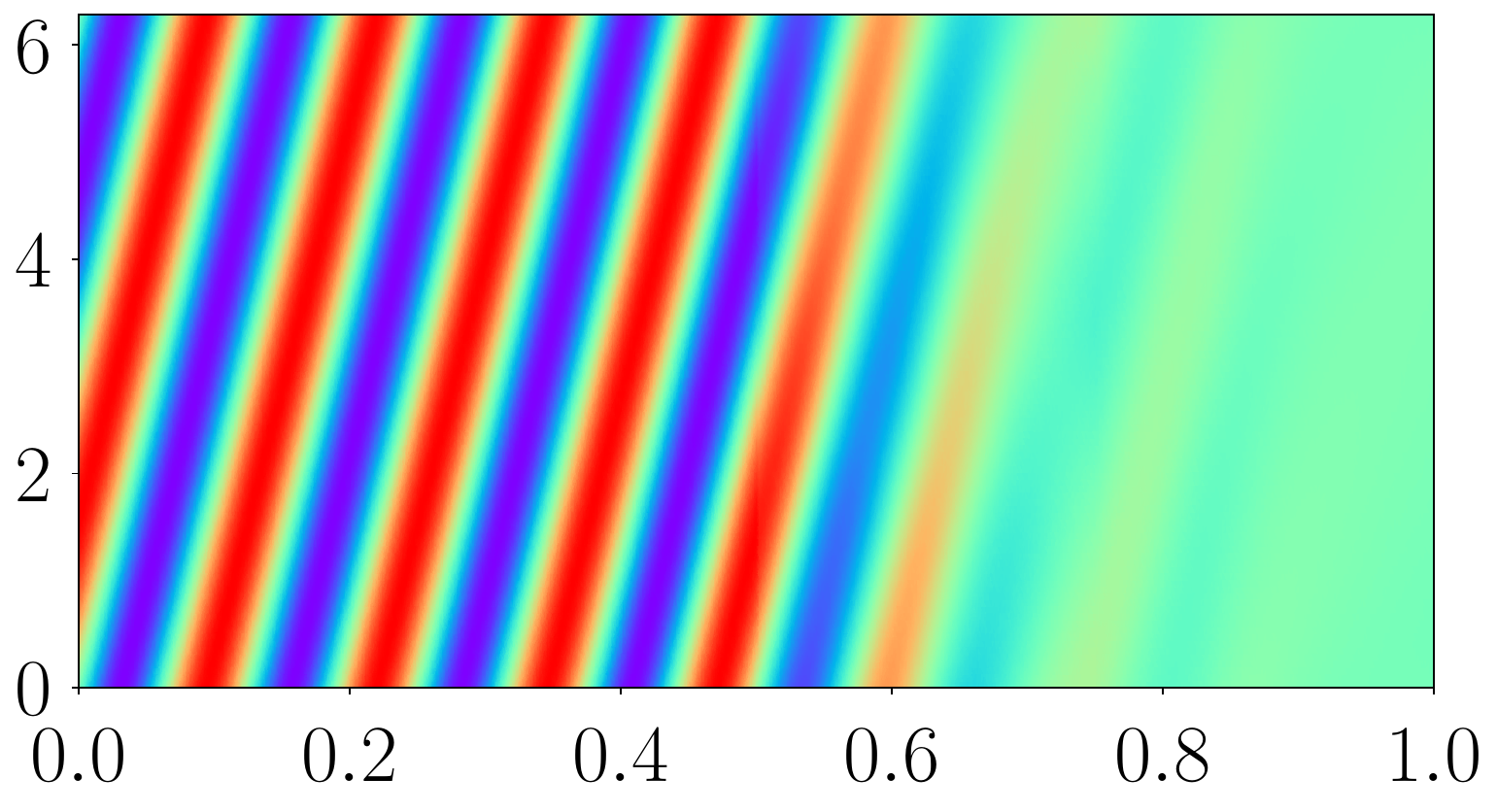} \\
  HCS-PINN, c = 100 &SCS-PINN, c = 100 \\
\includegraphics[trim={0.25cm 0 0.25cm 0},clip, width=0.46\linewidth]{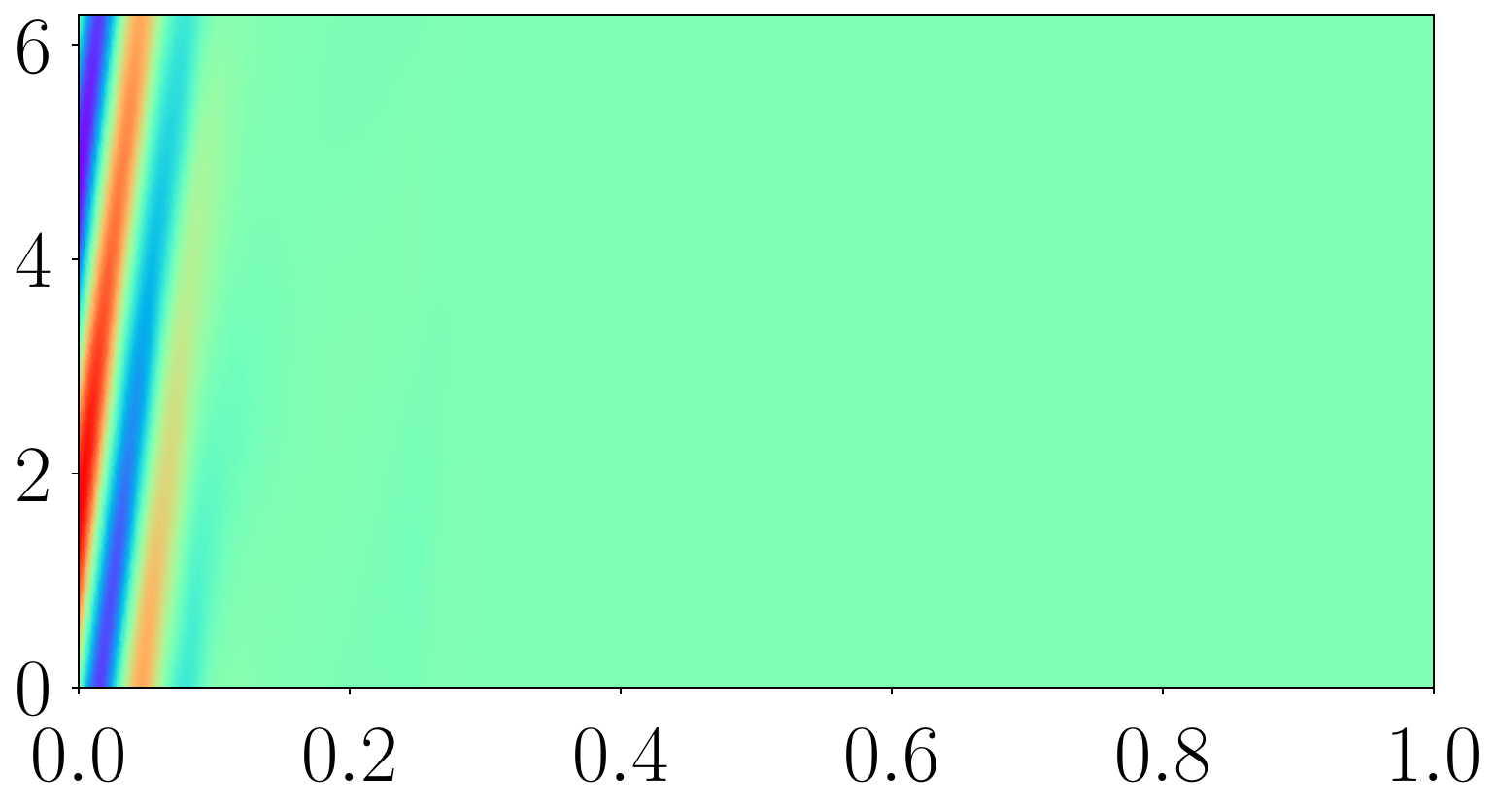}&
\includegraphics[trim={0.25cm 0 0.25cm 0},clip, width=0.46\linewidth]{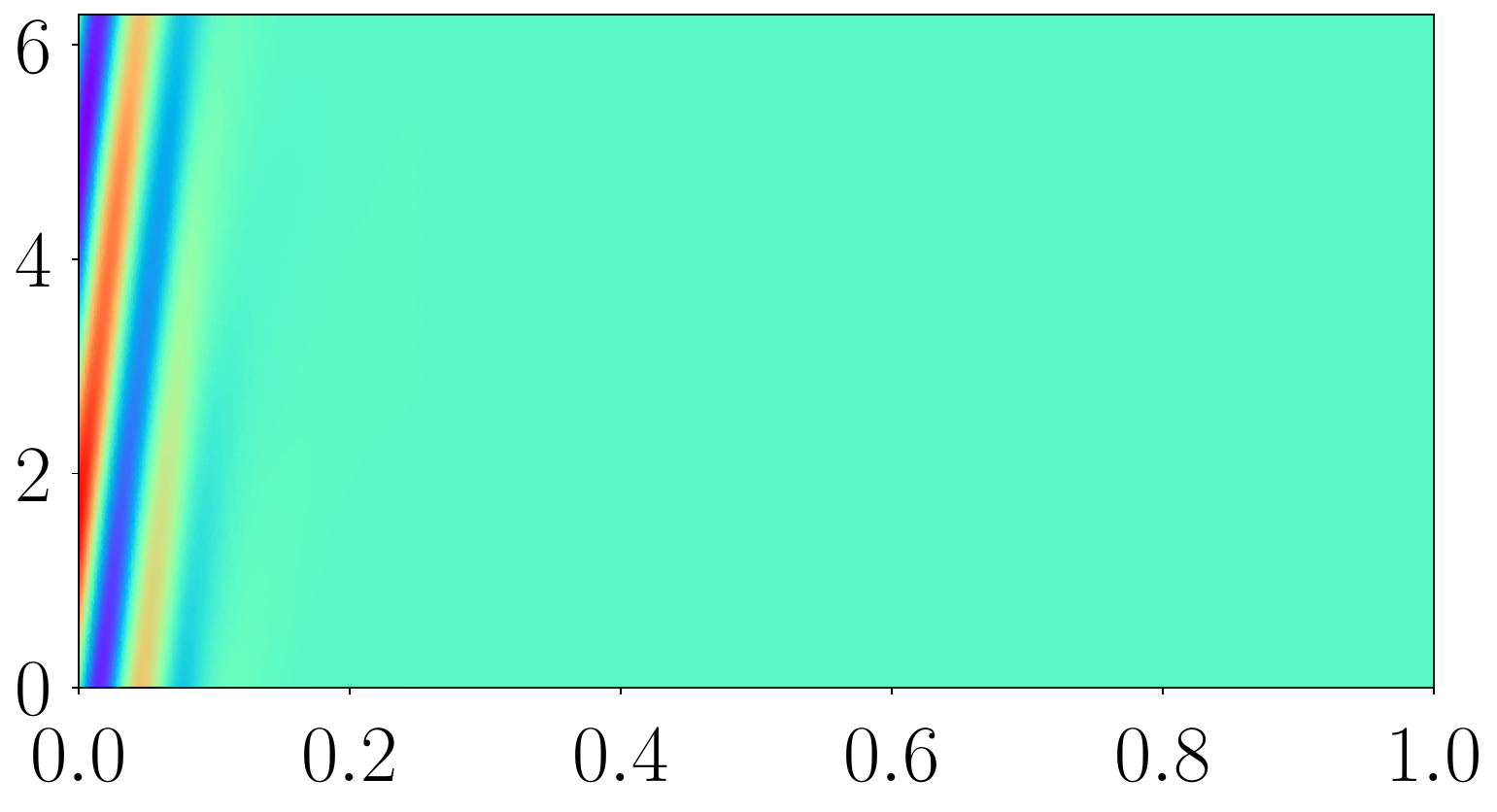} \\
\end{tabular}
\caption{HCS-PINN and SCS-PINN solution of advection equation for $nt = 4$.}
\label{fig:advection_nt4} 
\end{figure}

\begin{figure*}
   \centering
\begin{tabular}{cc}
  HCS-PINN, c = 30 &SCS-PINN, c = 30  \\ 
\includegraphics[trim={0.25cm 0 0.25cm 0},clip, width=0.46\linewidth]{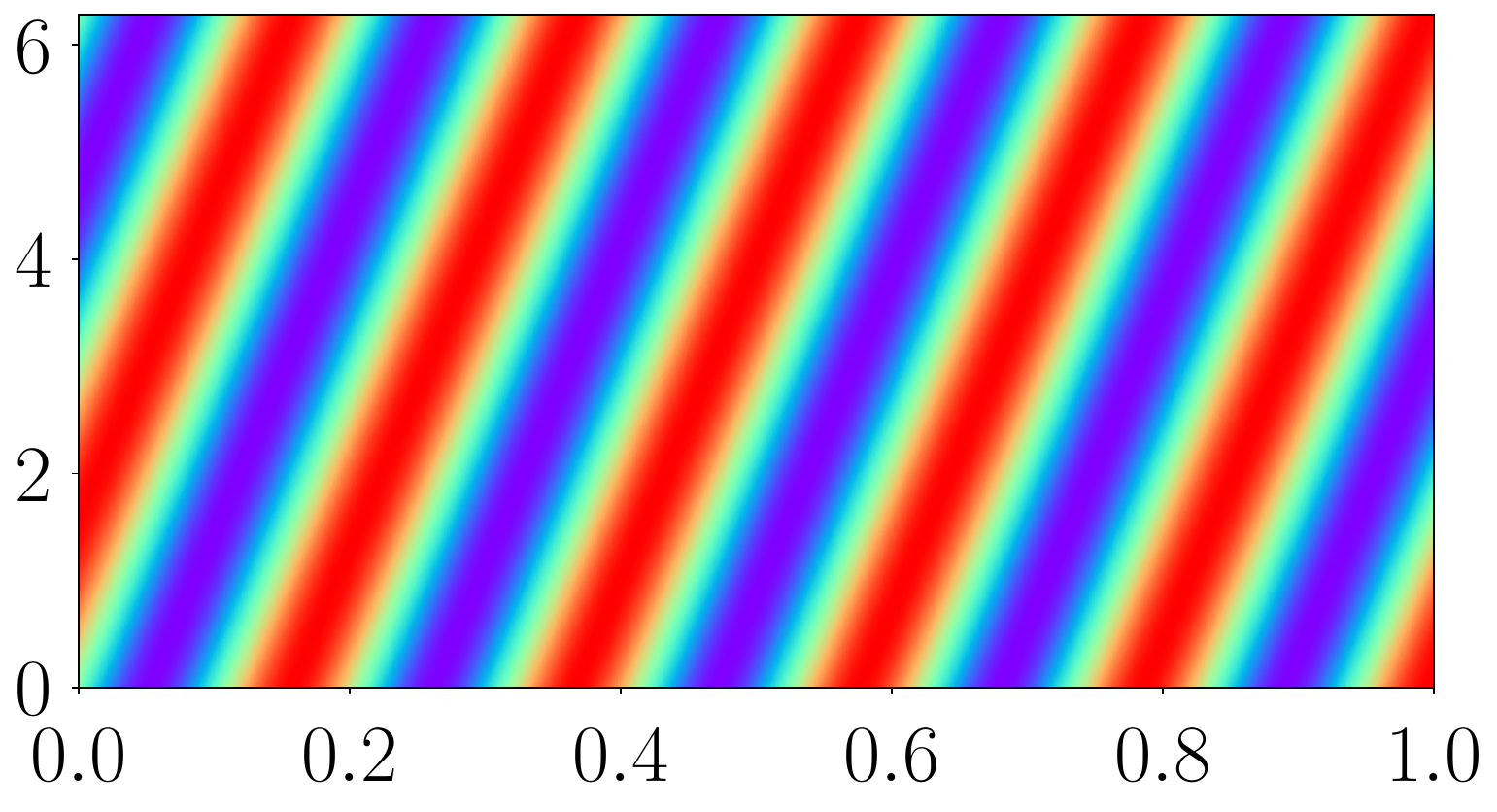}&
\includegraphics[trim={0.25cm 0 0.25cm 0},clip, width=0.46\linewidth]{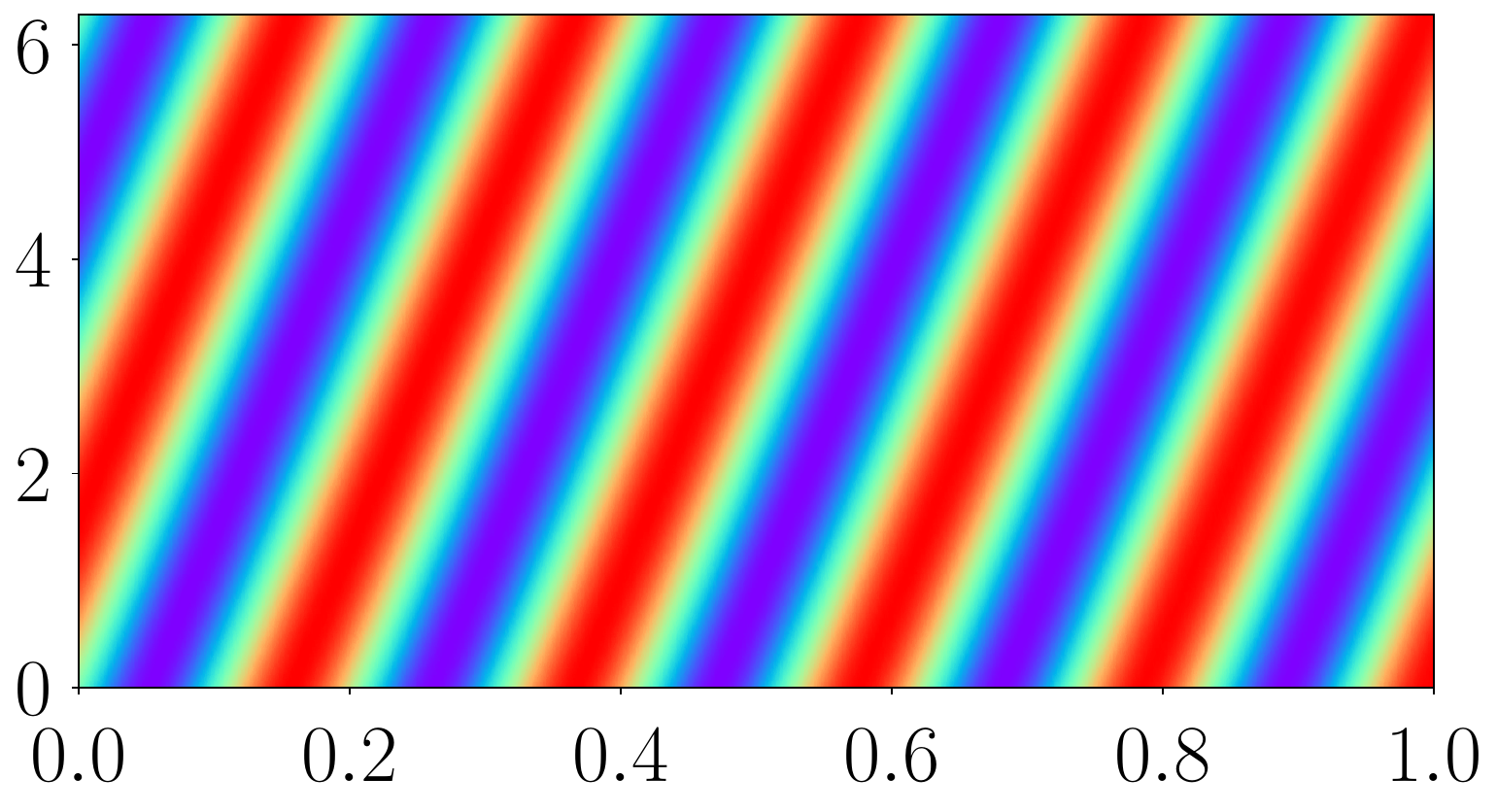}\\
 HCS-PINN, c = 50 &SCS-PINN, c = 50  \\
\includegraphics[trim={0.25cm 0 0.25cm 0},clip, width=0.46\linewidth]{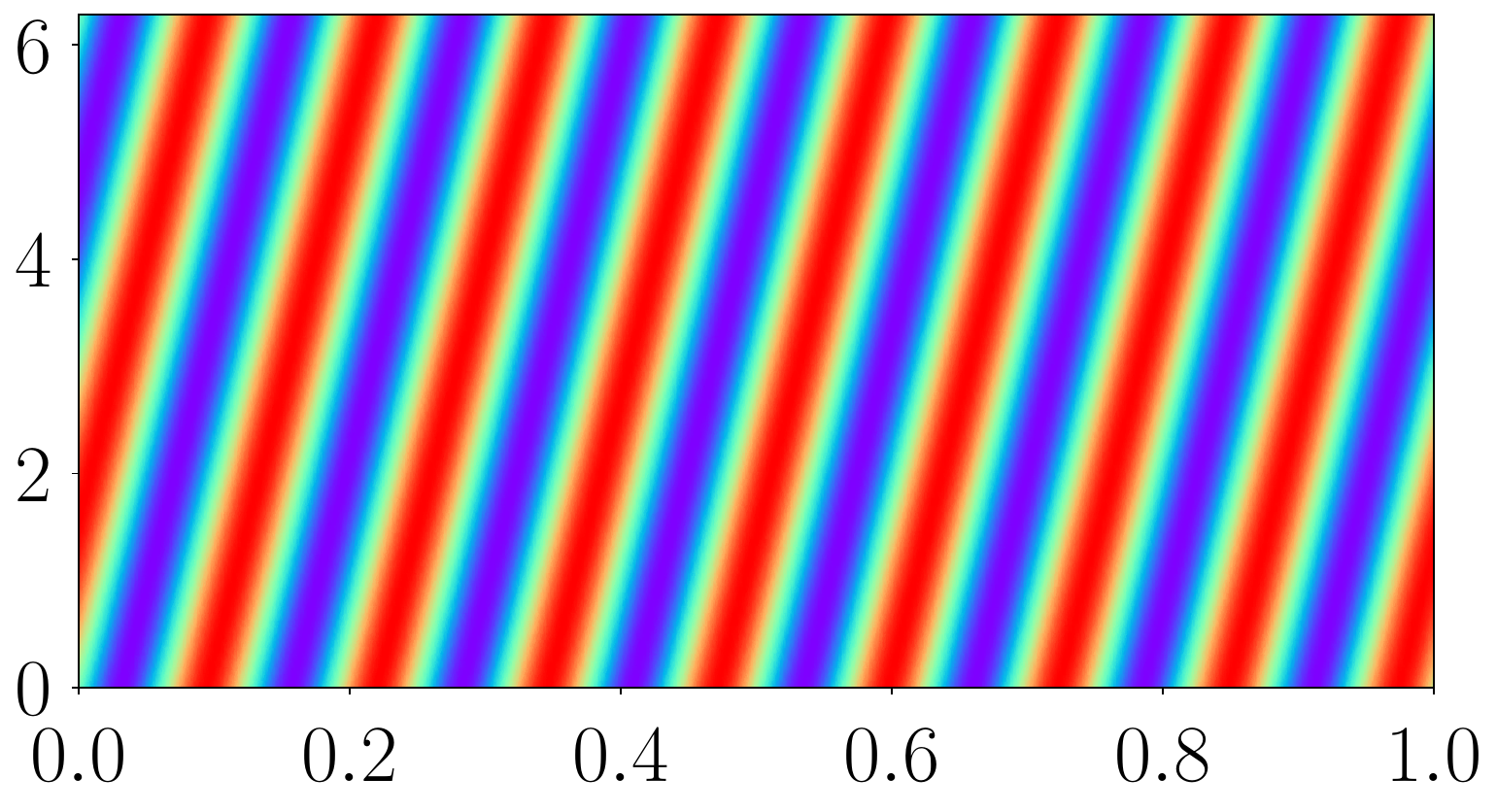}&
\includegraphics[trim={0.25cm 0 0.25cm 0},clip, width=0.46\linewidth]{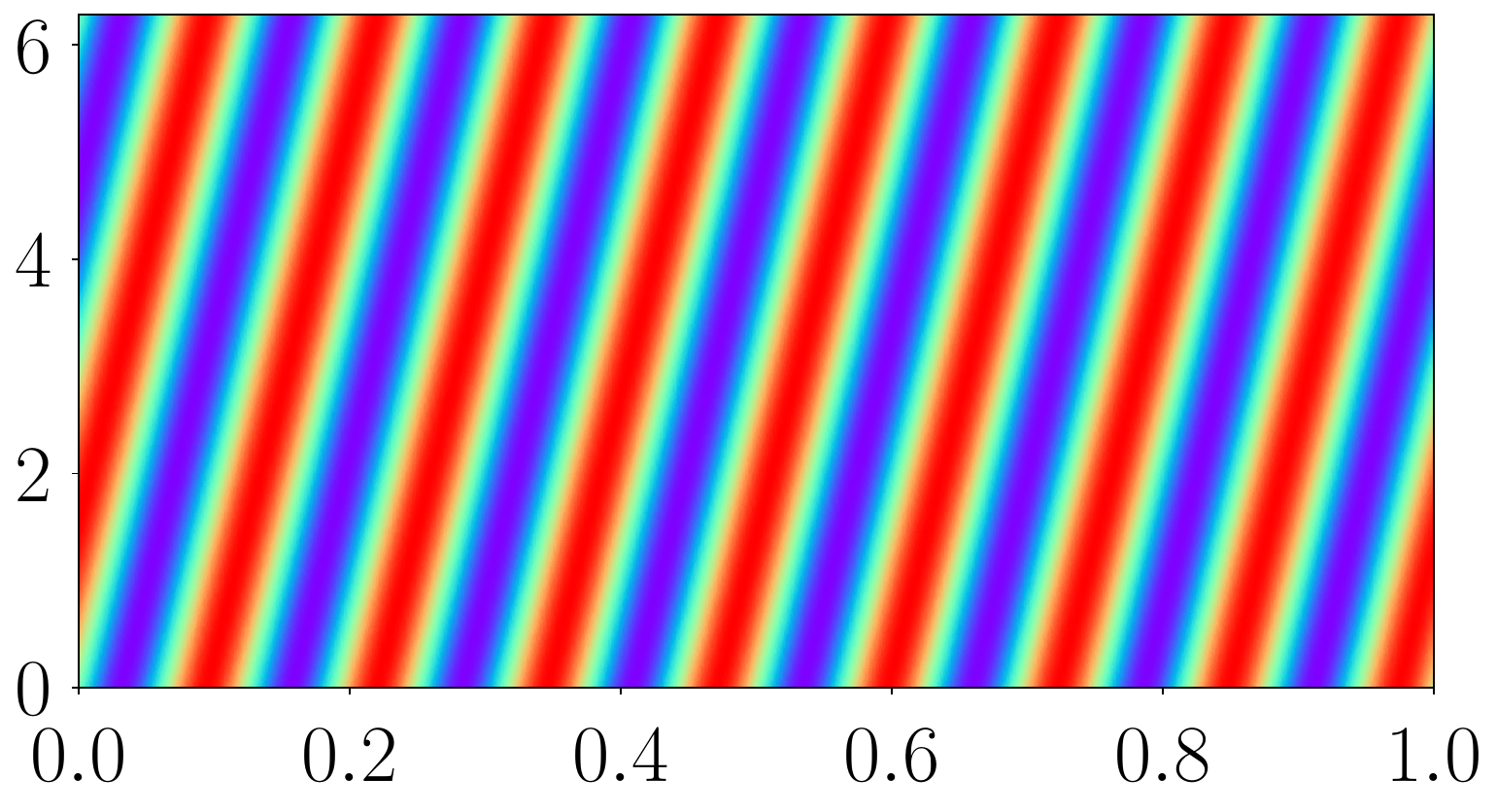} \\
  HCS-PINN, c = 100 &SCS-PINN, c = 100 \\
\includegraphics[trim={0.25cm 0 0.25cm 0},clip, width=0.46\linewidth]{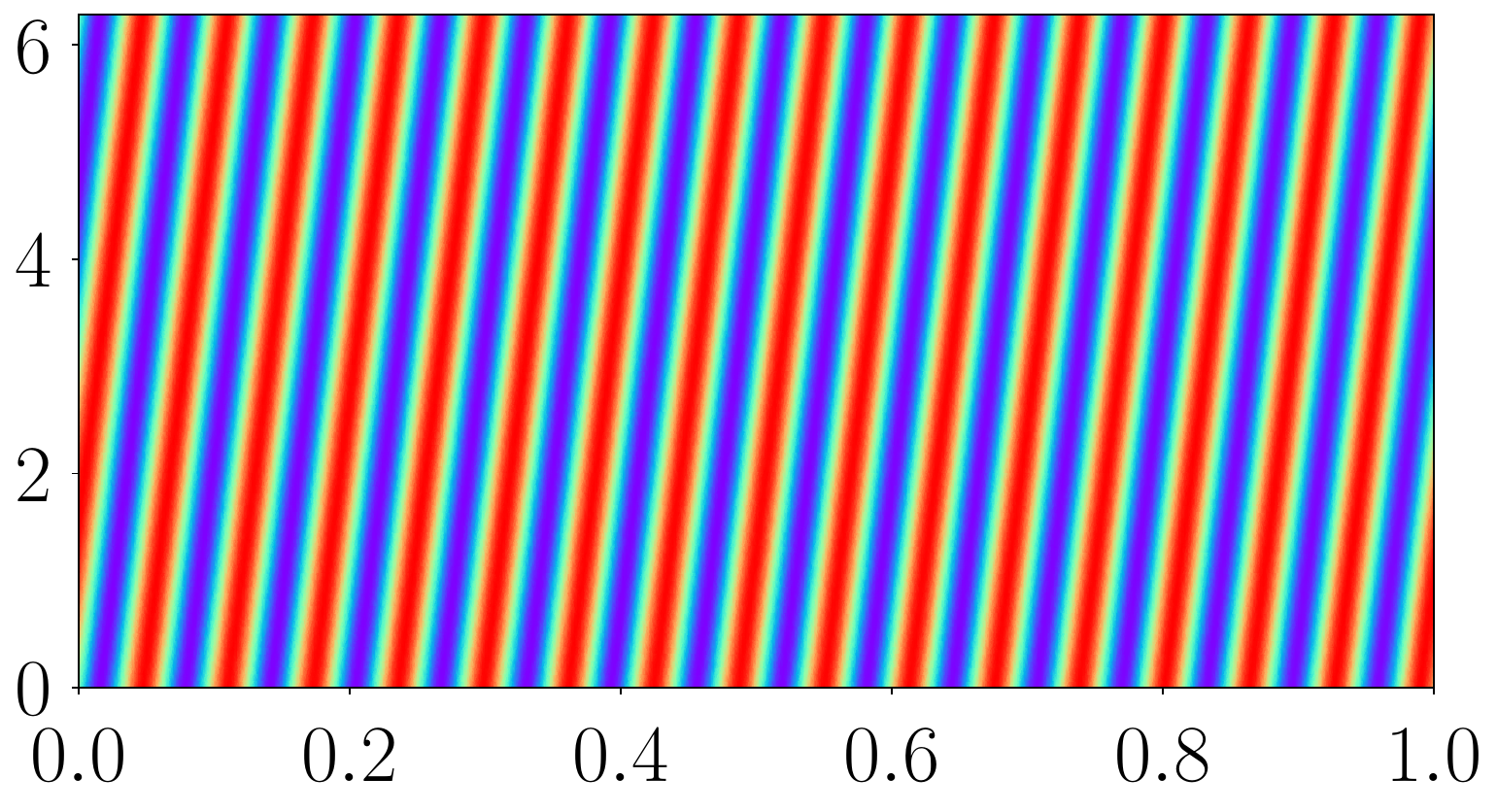}&
\includegraphics[trim={0.25cm 0 0.25cm 0},clip, width=0.46\linewidth]{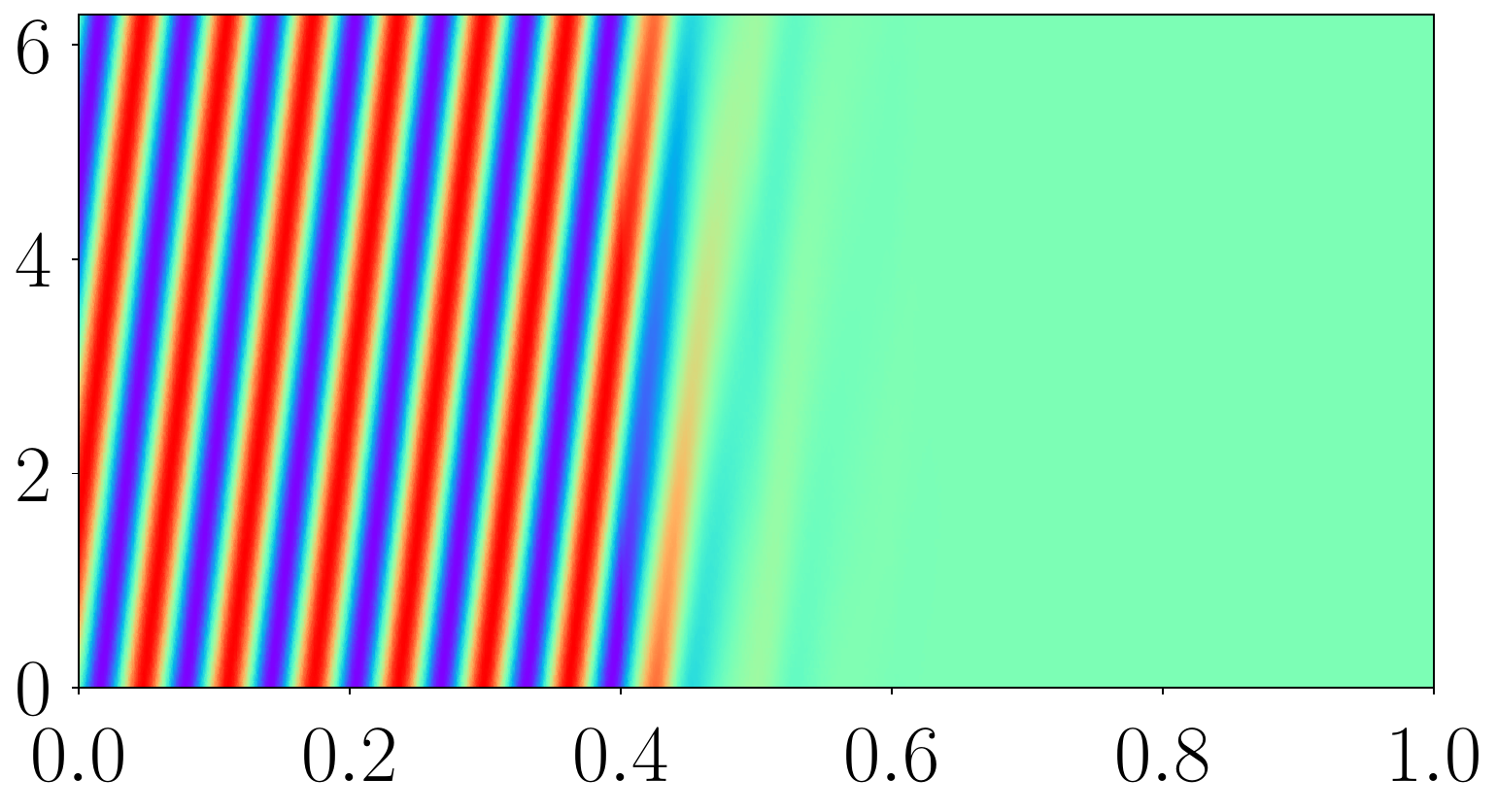} \\
\end{tabular}
\caption{HCS-PINN and SCS-PINN solution of advection equation for $nt = 10$.}
\label{fig:advection_nt10} 
\end{figure*}

When the fluid velocity is relatively large, vanilla PINNs struggle to capture the correct solution. In particular, PINN solutions often falsely converge to the zero solution, as seen in Figures \ref{fig:advection_nt4} and \ref{fig:advection_nt10}. For SCS-PINN, taking 10 time segments is sufficient to avoid this for $c=50$, but not for $c=100$. However, HCS-PINN captures the correct solution for all fluid speeds when taking 10 time steps, and captures up to $c=50$ when taking only 4.

Various other efforts have been made to mitigate this issue, including aforementioned temporal techniques as well as methods that encode characteristics into the architecture \cite{braga2022characteristics}. Wang et al. \cite{wang2023expert} suggested a range of techniques for obtaining optimal accuracy which include temporal and spatial Fourier feature embedding and grad norm weighting, among others. Indeed a relative $L^2$ error as low as O($10^{-4}$) was achieved for high transport velocity ($c=80$). In addition to the above techniques, they used a deep neural network (4 layers, 256 neurons in each layer, batch size of 8192) with a modified multi-layer perceptron (MLP). Here, we demonstrate that even with a relatively smaller network and a batch size (4 layers, 32 neurons in each layer, batch size of 128) using a standard MLP, a relative $L^2$ error of O($10^{-3}$) can be reached for high advection velocity cases (i.e. $c=100$) with HCS-PINN.

\begin{table}[H] 
\caption{Relative $L^2$ Error for advection velocity c = 30 (10,000 Adam iterations + 1,000 L-BFGS iterations, loss tolerance = $1 \times 10^{-6}$)}
\begin{center}
\begin{tabular}{ |c|c|c|c|c| } 
 \hline
 $nt$ & \thead{HCS-PINN} & \thead{HCS-PINN \\Time (secs)} & \thead{SCS-PINN} & \thead{SCS-PINN \\Time (secs)}\\  \hline
 1 & 8.6700  $\times$ 10$^{-1}$& 76 & 8.3647 $\times$ 10$^{-1}$& 47 \\ 
 4 & 3.4179 $\times$ 10$^{-3}$& 601 & 5.0206 $\times$ 10$^{-3}$ & 395\\ 
 10 & 2.1304 $\times$ 10$^{-3}$& 590 & 3.4125 $\times$ 10$^{-3}$ & 502 \\ 
 \hline
\end{tabular}
\end{center}
\end{table}

\begin{table}[H] 
\caption{Relative $L^2$ Error for advection velocity c = 50 (10,000 Adam iterations + 1,000 L-BFGS iterations, loss tolerance = $1 \times 10^{-6}$)}
\begin{center}
\begin{tabular}{ |c|c|c|c|c| } 
 \hline
 $nt$ & \thead{HCS-PINN} & \thead{HCS-PINN \\Time (secs)} & \thead{SCS-PINN} & \thead{SCS-PINN \\Time (secs)}\\  \hline
 1 & 9.5444  $\times$ 10$^{-1}$& 230 & 1.6658 $\times$ 10$^{0}$& 57 \\ 
 4 & 5.1898 $\times$ 10$^{-3}$& 1314 & 5.7477 $\times$ 10$^{-1}$ & 981 \\ 
 10 & 3.7520 $\times$ 10$^{-3}$& 1127 & 4.8585 $\times$ 10$^{-3}$ & 993 \\ 
 \hline
\end{tabular}
\end{center}
\end{table}

\begin{table}[H] 
\caption{Relative $L^2$ Error for advection velocity c = 100 (10,000 Adam iterations + 1,000 L-BFGS iterations, loss tolerance = $1 \times 10^{-6}$)}
\begin{center}
\begin{tabular}{ |c|c|c|c|c| } 
 \hline
 $nt$ & \thead{HCS-PINN} & \thead{HCS-PINN \\Time (secs)} & \thead{SCS-PINN} & \thead{SCS-PINN \\Time (secs)}\\  \hline
 1 & 1.02236  $\times$ 10$^{0}$& 56 & 9.9963 $\times$ 10$^{-1}$& 48 \\ 
 4 & 9.6380 $\times$ 10$^{-1}$& 619 & 9.7280 $\times$ 10$^{-1}$ & 146 \\ 
 10 & 6.2772 $\times$ 10$^{-3}$& 3812 & 7.3534 $\times$ 10$^{-1}$ & 1969 \\ 
 \hline
\end{tabular}
\end{center}
\end{table}

\subsection{Wave Equation} 
\begin{figure} [h]
	\centering
    \includegraphics[trim={0.25cm 0 0.25cm 0},clip, width=0.495\textwidth]{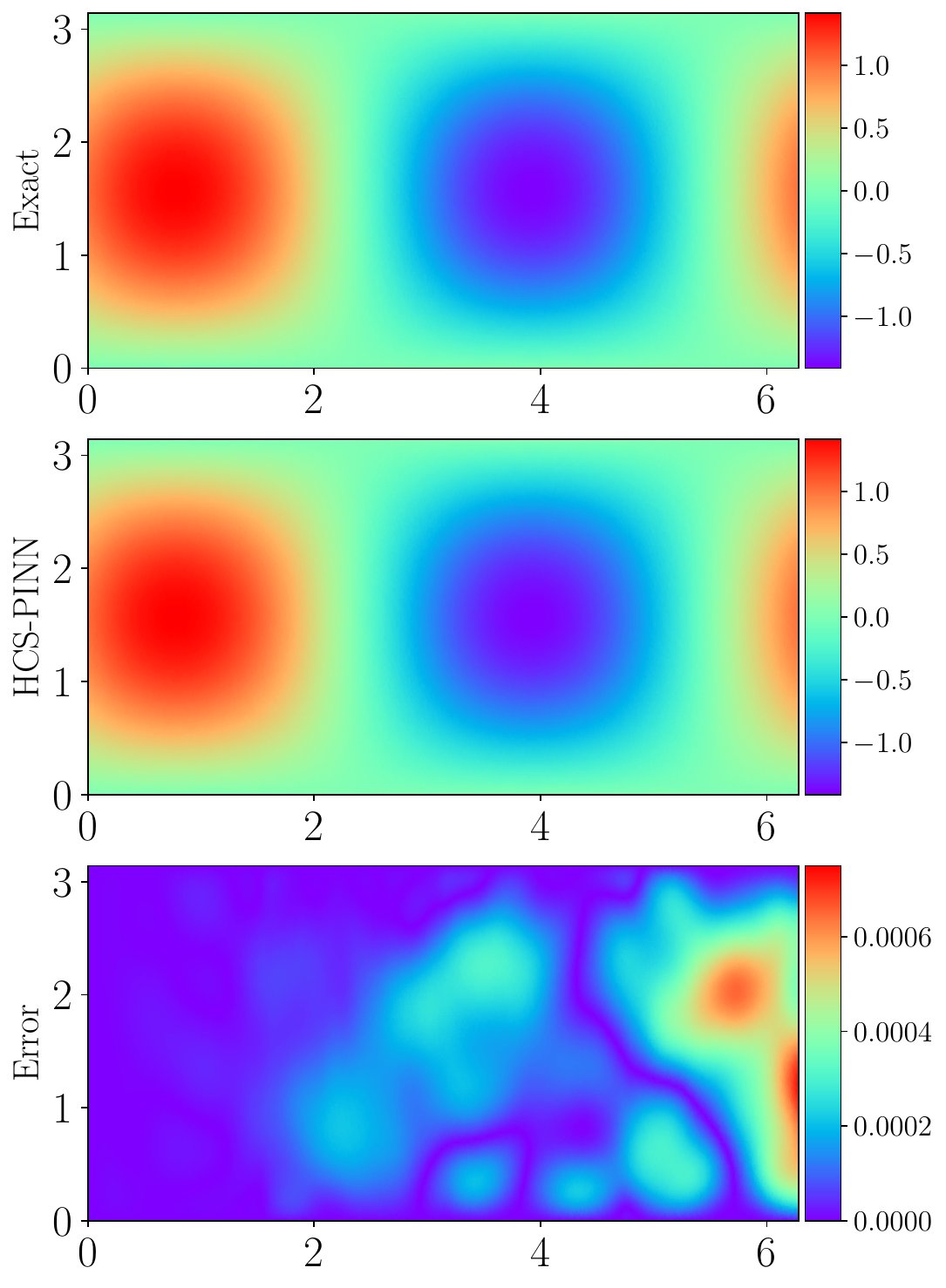}
    \includegraphics[trim={0.25cm 0 0.25cm 0},clip, width=0.495\textwidth]{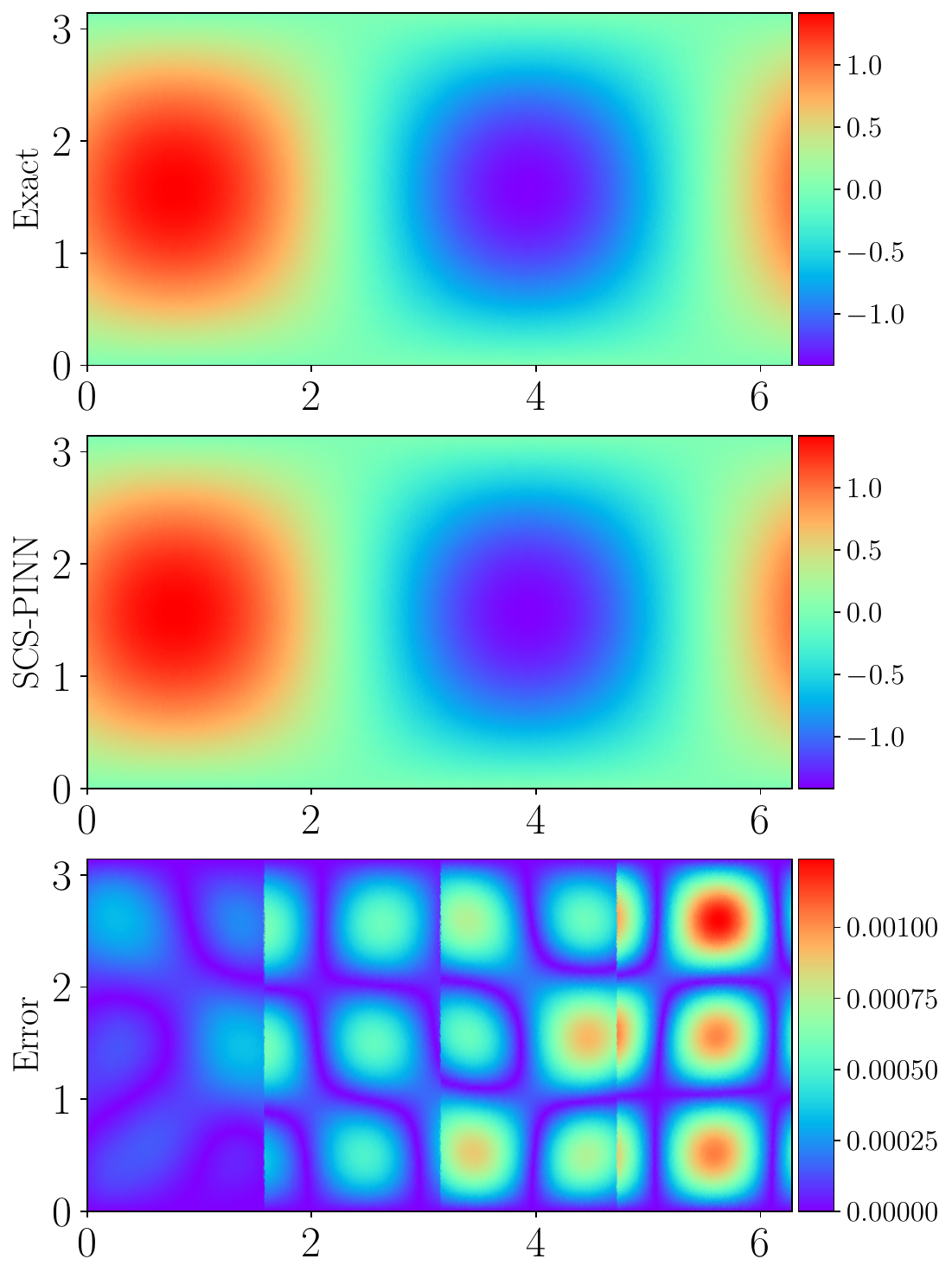}
	\caption{1D Wave equation contour for wave speed $c=1$ with HCS-PINN (left) and SCS-PINN (right). Even with 4 time segments, HCS-PINN produced a relative $L^2$ error of $2.6525\times 10^{-4}$. The corresponding SCS-PINN error is $5.1211\times 10^{-4}$.}
	\label{fig:wave_contour_c1}
\end{figure}

\begin{figure} [hp]
	\centering
    \includegraphics[trim={0.25cm 0 0.25cm 0},clip, width=0.495\textwidth]{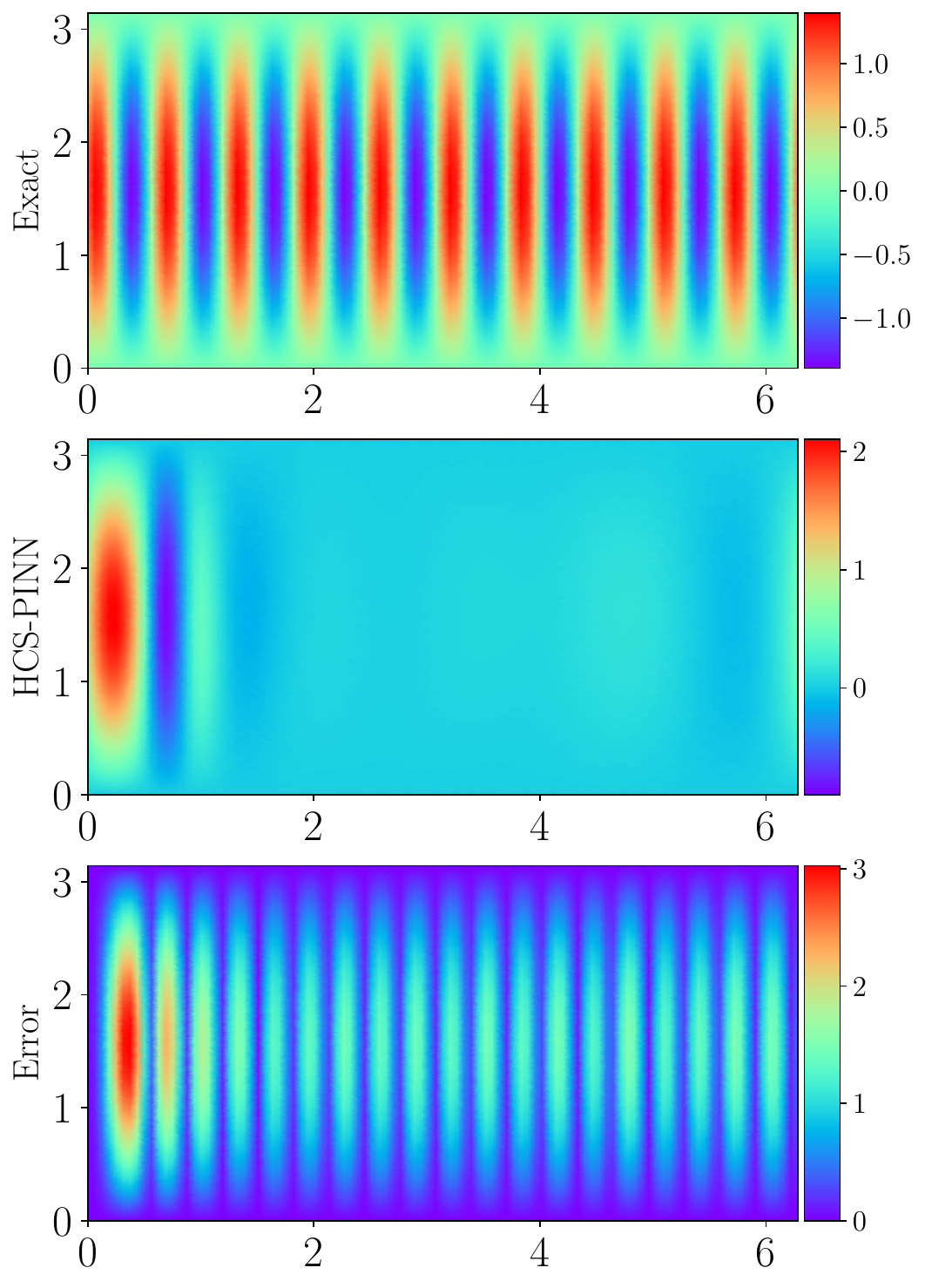}
	\includegraphics[trim={0.25cm 0 0.25cm 0},clip, width=0.495\textwidth]{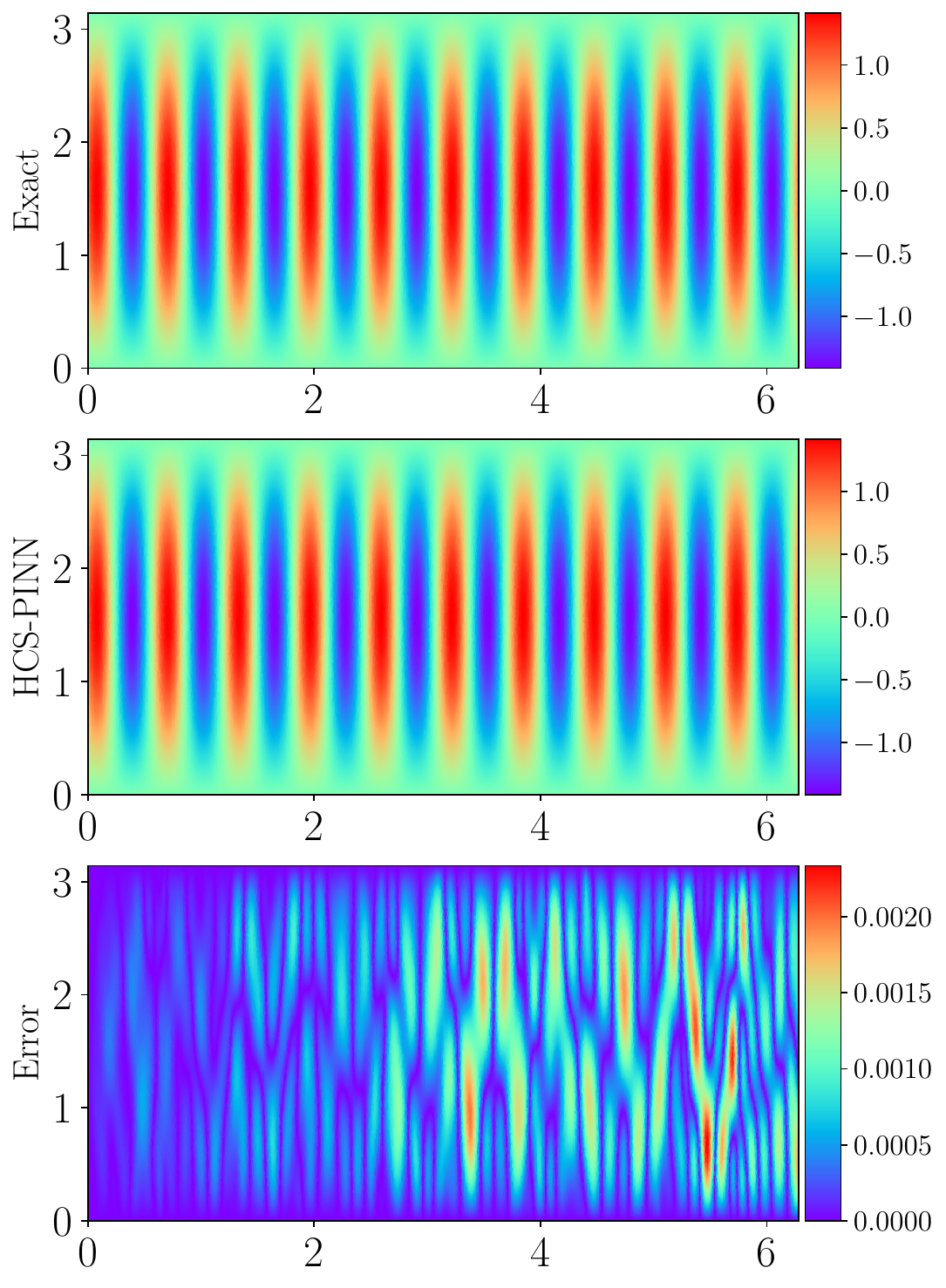}
	\caption{1D Wave equation contour for fluid velocity $c=10$ with HCS-PINN: $nt$ = 1(left), $nt = 10$ (right)}
	\label{fig:wave_contour_c10}
\end{figure}

Another hyperbolic system for which PINNs sometimes struggle is the following 1D wave equation:
\begin{align}
	u_{tt} &=c^2  u_{xx} ,\: x \in [0, \pi],\: t \in [0,2\pi] \\
	u(0,t) &= 0, \; u(\pi,t) = 0, \\
	u(x,0) &= sin(x), \; u_t(x,0) = c\:sin(x),
\end{align}
where $c$ is the wave propagation speed. The analytical solution of the above problem is given by: 
\begin{equation}
u(x,t) = sin(x)(sin(ct)+cos(ct))
\end{equation}
The second order derivative in time requires that we enforce $C^1$ continuity at time sub-domain interfaces. In addition to the temporal interpolation functions, the neural network is multiplied by the term $x(\pi -x)$ to strongly enforce zero Dirichlet boundary conditions. Similar to the advection equation, the PDE was scaled by the square of the wave speed in lieu of altering the weighting parameter $\lambda_I$ for SCS-PINNs. For low wave speeds, such as $c=1$ or less, it is relatively easy to get accurate solutions (see Figure \ref{fig:wave_contour_c1}). However, as the wave speed increases, this becomes increasingly difficult with vanilla PINNs. For example, when $c=10$, neither the HCS-PINN or SCS-PINN converges when only a single time segment is used, as shown in Figure~\ref{fig:wave_contour_c10}. For the cases of 4 and 10 time windows, HCS-PINN gives less error than the corresponding SCS-PINN. Additionally, the 10 time step case actually takes less time than the 4 time step case, as it continually reaches the tolerance criterion earlier. These results are tabulated in Table \ref{tab:wave_c10}.   

While the SCS-PINN solution may seem continuous over time, the presence of discontinuities becomes evident in the error contour depicted in Figure \ref{fig:wave_contour_c1}. 
For both low and high fluid velocities $c=1$ and $c=10$, relative $L^2$ errors of O($10^{-4}$) were achieved using the HCS-PINN method. 

\begin{table}[h]
\caption{Relative $L^2$ Error for wave equation with wave speed $c=10$ (20,000 iterations of Adam + 1,000 L-BFGS iterations, loss tolerance = $1 \times 10^{-6}$)}
\begin{center}
\begin{tabular}{ |c|c|c|c|c| } 
 \hline
 $nt$ & \thead{HCS-PINN} & \thead{HCS-PINN \\Time (secs)} & \thead{SCS-PINN} & \thead{SCS-PINN \\Time (secs)}\\  \hline
 1 & 1.1642 $\times$ 10$^{0}$& 718 & 9.8927 $\times$ 10$^{-1}$& 598 \\ 
 4 & 1.5629 $\times$ 10$^{-3}$& 2760 & 3.1508 $\times$ 10$^{-2}$ & 3564\\ 
 10 & 8.2517 $\times$ 10$^{-4}$& 2188 & 1.9872 $\times$ 10$^{-3}$ & 7348 \\ 
 \hline
\end{tabular}
\end{center}
\label{tab:wave_c10}
\end{table}

\subsection{Allen-Cahn Equation} 
The Allen-Cahn equation is used for the evolution of phase transitions as well as more general interface motion \cite{allen1979microscopic}. A particular form used in benchmarking PINNs is given by
\begin{align}
	&u_t - \lambda_1 u_{xx} + \lambda_2 u^3 - \lambda_2 u  =0,\:  x \in [-1, 1],\: t \in [0,1] \\
	&u(0,x) = x^2 cos(\pi x) \\
	&u(t,-1) = u(t,1), \, u_x(t,-1) = u_x(t,1)
\end{align}
where $\lambda_1=0.0001$ and $\lambda_2=5$. The Allen-Cahn equation requires $C^0$ continuity between sequential neural networks. 
The interplay between the diffusion term and non-linear source terms makes it a particularly challenging problem, posing difficulties even for traditional numerical methods. The stiffness of the equation is dictated by the relative magnitude of the diffusion and reaction coefficients, i.e. $\lambda_1$ and $\lambda_2$, respectively. A number of methods, including Adaptive Time Sampling and bc-PINNs, were developed with this specific problem in mind. 

Here, all cases ($nt=1,4,10$) converge as indicated by Table \ref{Table:AC_adam_lbfgs}, with HCS-PINN reaching a relative $L^2$ error of O($10^{-4}$) and SCS-PINN reaching O($10^{-3})$. While both methods take approximately the same amount of time in these simulations, HCS-PINN produced more accurate results for a given number of time steps. Note that we have used a penalty of 100 for $\lambda_I$ in the SCS-PINN loss terms associated with temporal continuity and initial conditions. Without this penalty, the accuracy of SCS-PINN was even lower. 
\begin{figure} [H]
	\centering
    \includegraphics[trim={0.25cm 0 0.25cm 0},clip, width=0.495\textwidth]{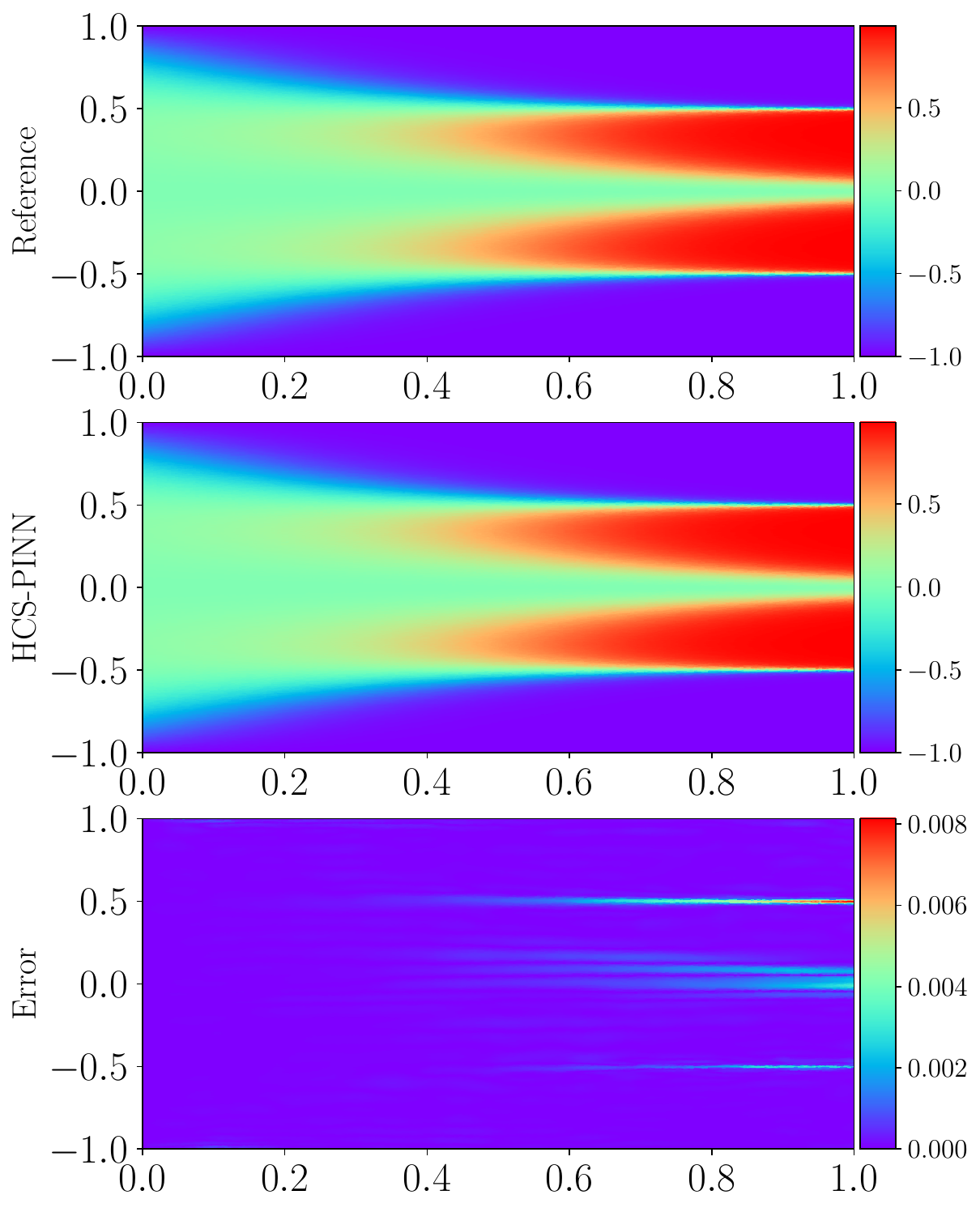}
	\includegraphics[trim={0.25cm 0 0.25cm 0},clip, width=0.495\textwidth]{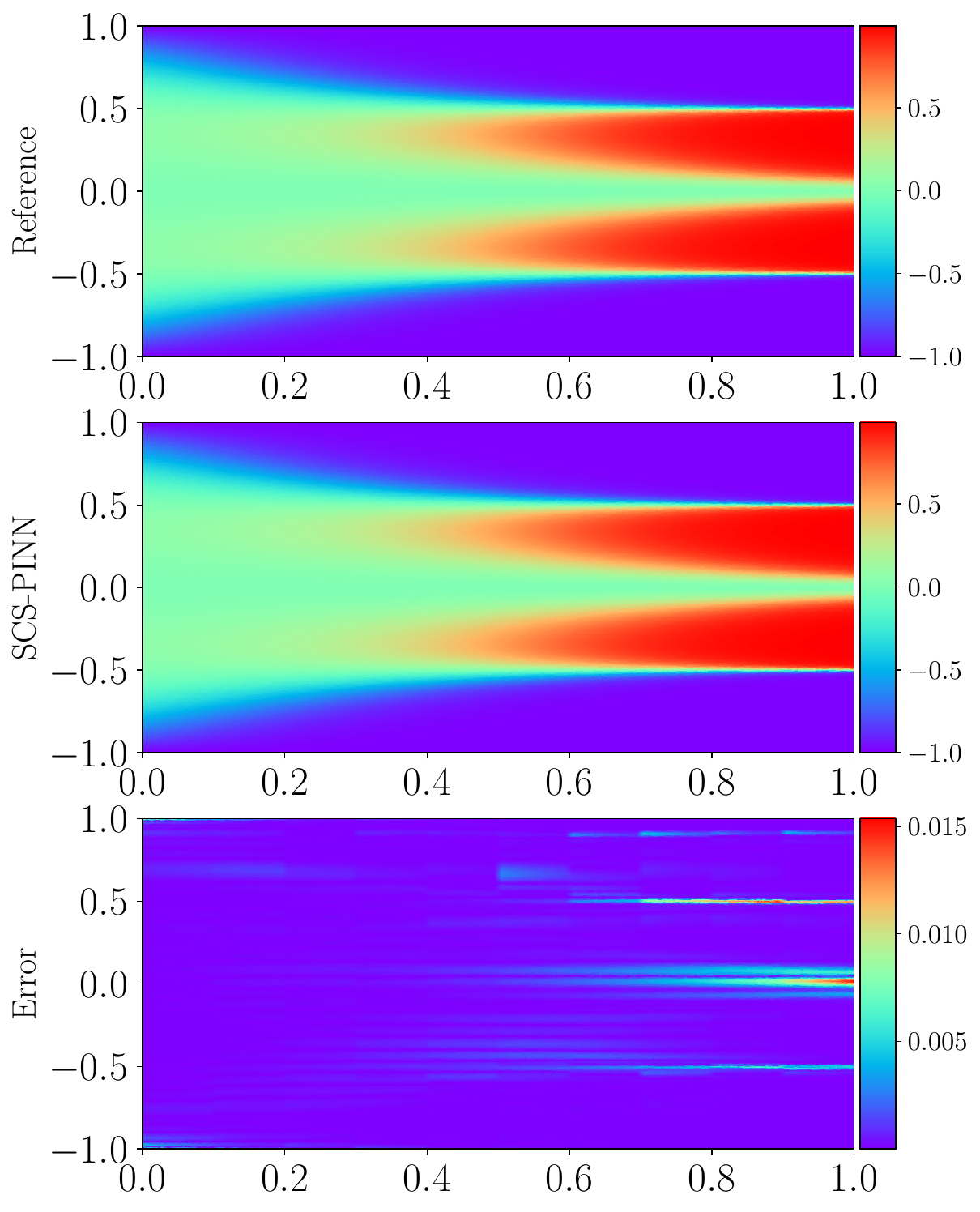}
	\caption{Contours of solution and errors of Allen-Cahn equation for  $nt=10$ with HCS-PINN (left), and SCS-PINN (right).}
	\label{fig:AC_contour}
\end{figure}

\begin{table}[h] 
\caption{Relative $L^2$ Error for Allen-Cahn equation (20,000 Adam iterations + 200 L-BFGS iterations, batch size = 256, loss tolerance = $1 \times 10^{-7}$)}
\begin{center}
\begin{tabular}{ |c|c|c|c|c| } 
 \hline
 $nt$ & \thead{HCS-PINN} & \thead{HCS-PINN \\Time (secs)} & \thead{SCS-PINN} & \thead{SCS-PINN \\Time (secs)}\\  \hline
 1 & 1.0455 $\times$ 10$^{-2}$& 348 & 4.3057 $\times$ 10$^{-2}$& 368 \\ 
 4 & 1.0650 $\times$ 10$^{-3}$& 1495 & 2.8939 $\times$ 10$^{-3}$ &  1426\\ 
 10 & 5.4117 $\times$ 10$^{-4}$& 3222 & 1.4205 $\times$ 10$^{-3}$ & 3639 \\ 
 \hline
\end{tabular}
\end{center}
\label{Table:AC_adam_lbfgs}
\end{table}

\begin{figure} [H]
	\centering
	\includegraphics[width=\textwidth]{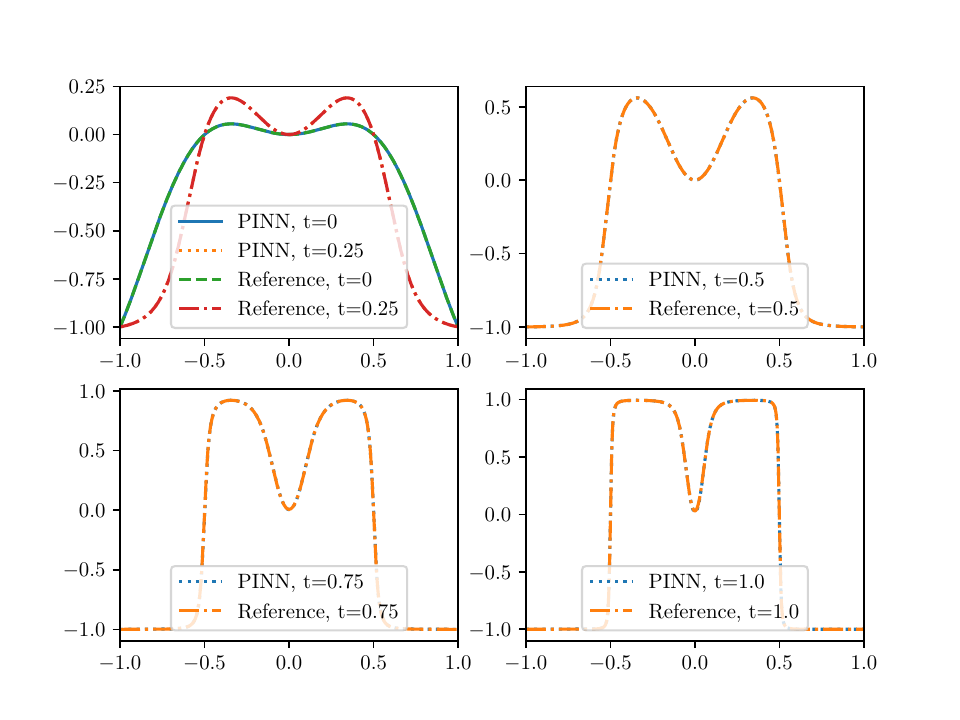}
	\caption{Comparison of Allen-Cahn equation with reference solution \cite{driscoll2014chebfun} for different time snapshots with HCSPINN ($nt = 4$).}
	\label{fig:ac_comparison}
\end{figure}

\subsection{Korteweg–de Vries (KdV) Equation} 
The Korteweg–de Vries (KdV) equation, which depicts the evolution of waves on shallow water surfaces \cite{miura1976korteweg}, is another common benchmark problem for PINNs. The particular problem is given by
\begin{align}
	&u_t + \lambda_1 u u_{x} + \lambda_2 u_{xxx} = 0, \: x \in [-1,1],\: t \in [0, 1] \\
	&u(0,x) = cos(\pi x) \\
	&u(t,-1) = u(t,1), \, u_x(t,-1) = u_x(t,1)
\end{align}
where $\lambda_1= 1$ and $\lambda_2= 0.0025$. The KdV equation contains three distinct terms. The unsteady term $u_t$ requires $C^0$ continuity, while the $uu_x$ term is a non-linear advection term and $u_{xxx}$ is a dispersion term. The initial cosine wave therefore undergoes both advection and dispersion as it evolves, resulting in the generation of higher frequency modes over time. This evolution poses a challenge for PINN algorithms in accurately predicting the long-term behavior of the KdV equation.   

\begin{figure} [H]
	\centering
    \includegraphics[trim={0.25cm 0 0.25cm 0},clip, width=0.495\textwidth]{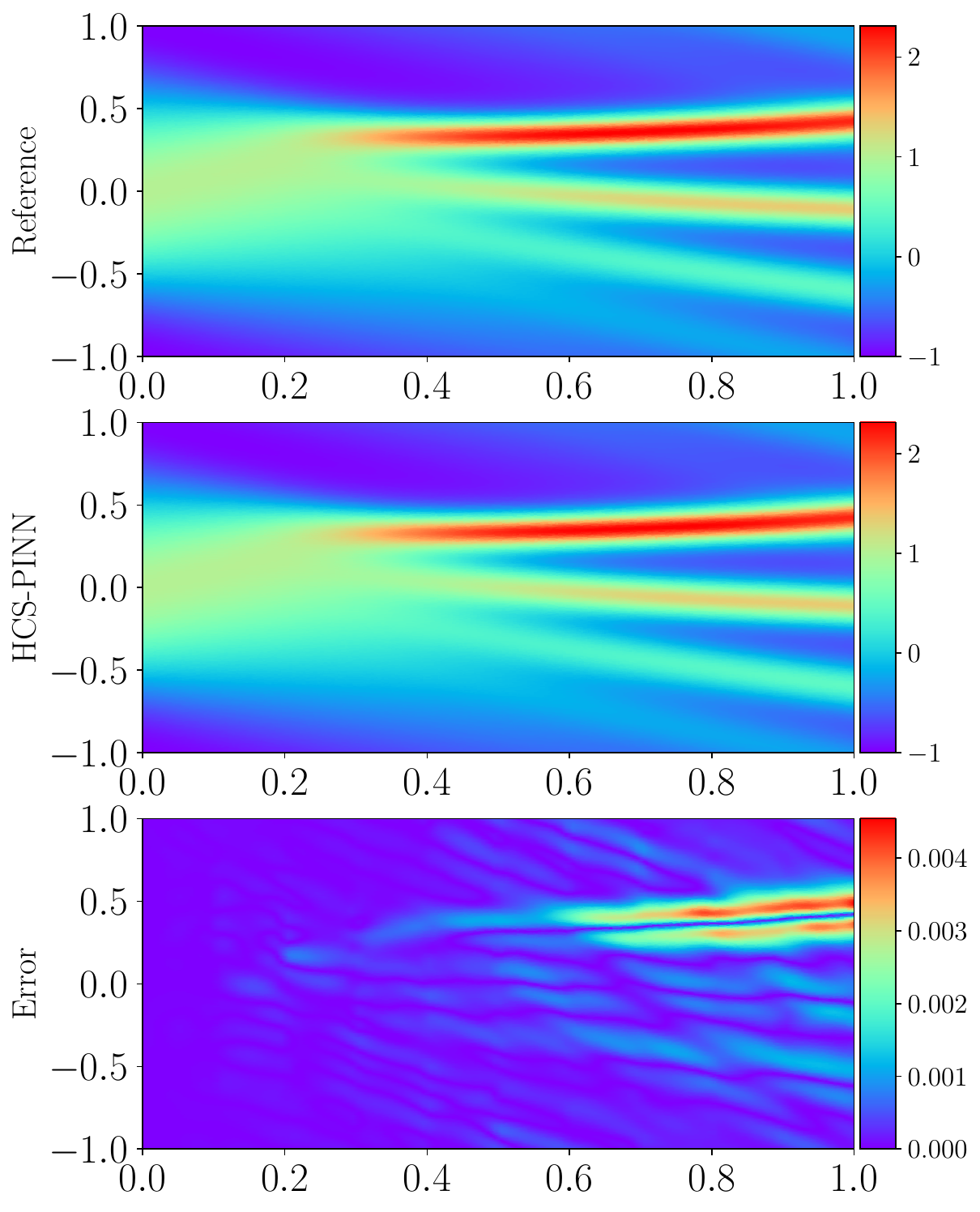}
	\includegraphics[trim={0.25cm 0 0.25cm 0},clip, width=0.495\textwidth]{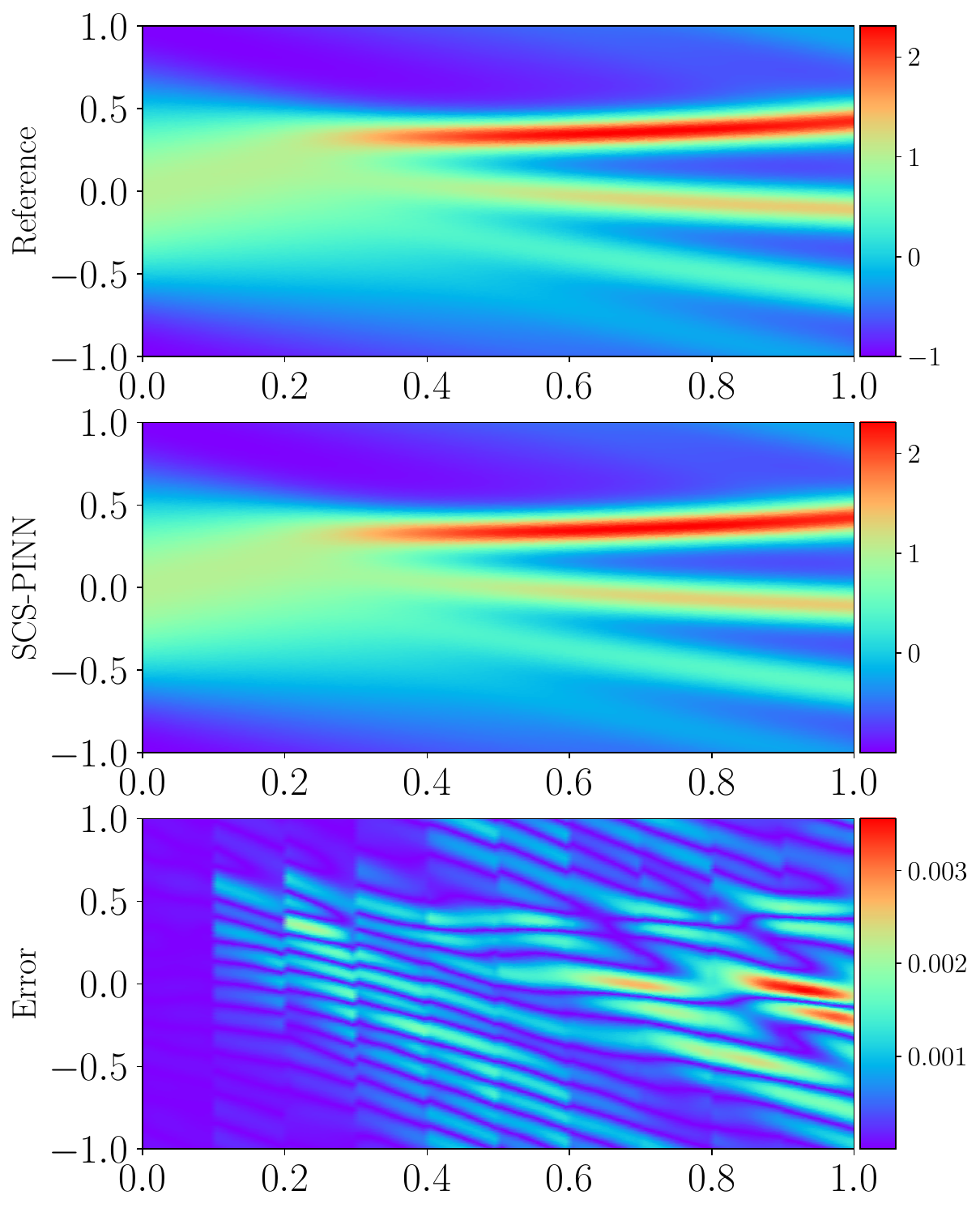}
	\caption{Contours of solution and errors of KdV equation for  $nt=10$ with HCS-PINN (left), and SCS-PINN (right).}
	\label{fig:KdV_contour}
\end{figure}

\begin{figure} [H]
	\centering
	\includegraphics[width=\textwidth]{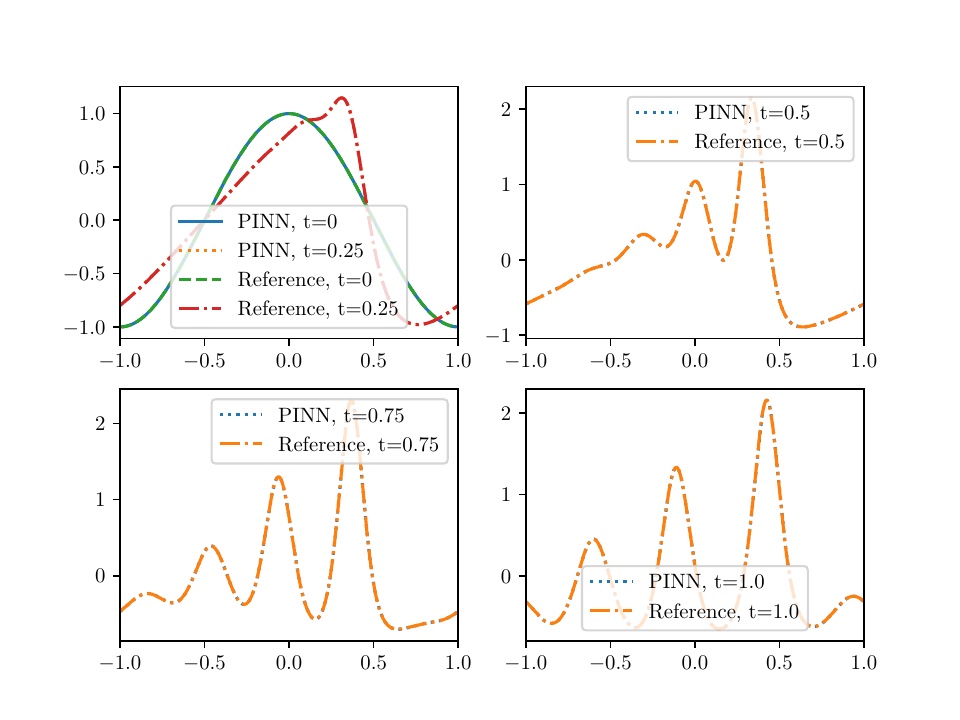}
	\caption{Comparison of KdV equation with reference solution \cite{driscoll2014chebfun} for different time snapshots with HCS-PINN ($nt=4$).}
	\label{fig:kdv_comparison}
\end{figure}

The solution is well captured by the SCS-PINN and especially the HCS-PINN within the time range [0,1], as shown in Figures \ref{fig:KdV_contour} and \ref{fig:kdv_comparison}. Again, the discontinuities at the time window interfaces in the error contour for SCS-PINN makes apparent the soft nature of continuity enforcement. 

A comparison of the errors from the two approaches is listed in Table \ref{tab:KdV}. As with the Allen-Cahn equation, the HCS-PINN shows better accuracy than the SCS-PINN using the same amount of training time. Again, we have also tuned the penalty parameter $\lambda_I$ to 100 to improve the SCS-PINN solution. 

\begin{table}[h] 
\caption{Relative $L^2$ Error for KdV equation (20,000 Adam iterations + 200 L-BFGS iterations, loss tolerance = $1 \times 10^{-6}$)}
\begin{center}
\begin{tabular}{ |c|c|c|c|c| } 
 \hline
 $nt$ & \thead{HCS-PINN} & \thead{HCS-PINN \\Time (secs)} & \thead{SCS-PINN} & \thead{SCS-PINN \\Time (secs)}\\  \hline
 1 & 1.1506 $\times$ 10$^{-2}$& 701 & 3.2753 $\times$ 10$^{-2}$& 684 \\ 
 4 & 1.3264 $\times$ 10$^{-3}$& 2764 & 2.6425 $\times$ 10$^{-3}$ & 2738\\ 
 10 & 9.7535 $\times$ 10$^{-4}$& 7128 & 1.0520 $\times$ 10$^{-3}$ & 6477 \\ 
 \hline
\end{tabular}
\end{center}
\label{tab:KdV}
\end{table}

\subsection{ Chaotic Dynamics }
Many dynamical systems which exhibit chaotic behavior, such as the famous Lorenz equation~\cite{lorenz1963deterministic}, are cast in terms of three first order autonomous ODEs, i.e. 
 \begin{equation}
   \begin{split}
       x_{t}&=f_1(x,y,z) \\  
       y_{t}&=f_2(x,y,z) \\  
       z_{t}&=f_3(x,y,z) \\  .
   \end{split}
 \end{equation}
In fact, this is a minimal setting for chaotic solutions, as one or two first order continuous autonomous equations cannot produce chaos according to Poincare-Bendixon theorem~\cite{smale1974differential}. 
In the years that followed Lorenz's pioneering work, various efforts were made to develop simple chaotic models of this form, of which Rossler~\cite{rossler1979chaos} and Sprott \cite{sprott1997some} were at the forefront. 
Many of these models can be recast into a single, third order Jerk equation of the form:
\begin{equation}
    x_{ttt}=J(x,x_{t},x_{tt}) 
\end{equation}
In particular, Eichhorn et al.~\cite{eichhorn1998transformations} systematically showed that Rossler's prototype-4 model and many of Sprott's equations can undergo this type of transformation. 
Additionally, they outlined seven different classes of dissipative, chaotic Jerk systems. One of the simplest is the $JD_2$ class, which is given by  

\begin{equation}
    x_{ttt} =k_1 x_{tt} +k_2 x_{t} + x^2 +k_3 \, .
\end{equation}
This is used here as an example problem, with $k_1=-0.4$, $k_2=-2.1$, and $k_3=-1.0$ as suggested by Sprott~\cite{sprott2009simplifications}. 
Additionally we take $x(0)=0$, $x_{t}(0)=1.0$, $x_{tt}(0)=1.0$ as initial conditions.

\begin{figure*}
   \centering
\begin{tabular}{cc}
  HCS-PINN, $nt = 5$ &SCS-PINN, $nt = 5$  \\ 
\includegraphics[width=0.46\linewidth]{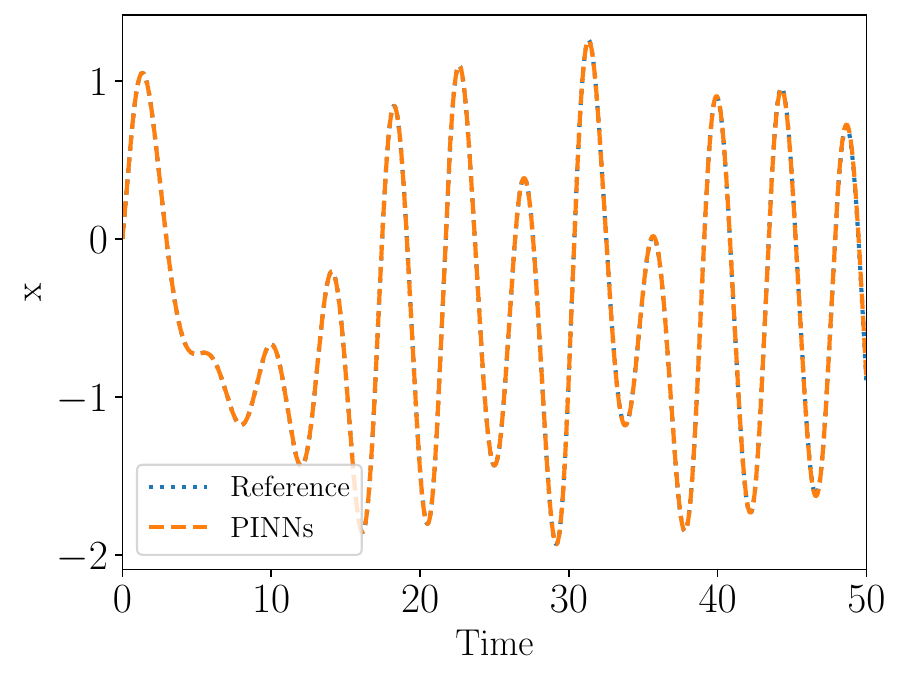}&
\includegraphics[width=0.46\linewidth]{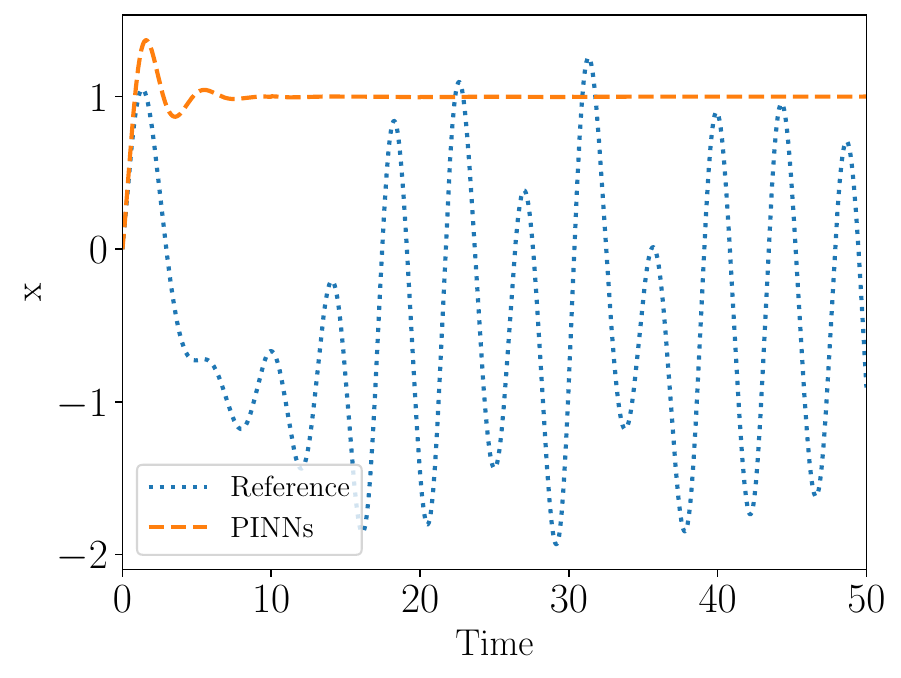}\\
 HCS-PINN, $nt = 10$ &SCS-PINN, $nt = 10$  \\
\includegraphics[width=0.46\linewidth]{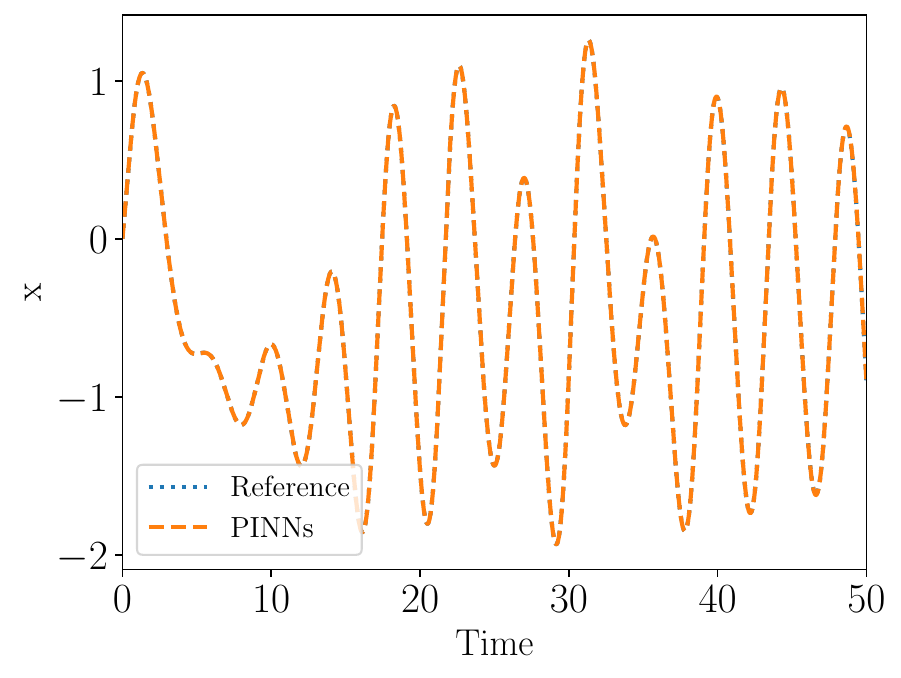}&
\includegraphics[width=0.46\linewidth]{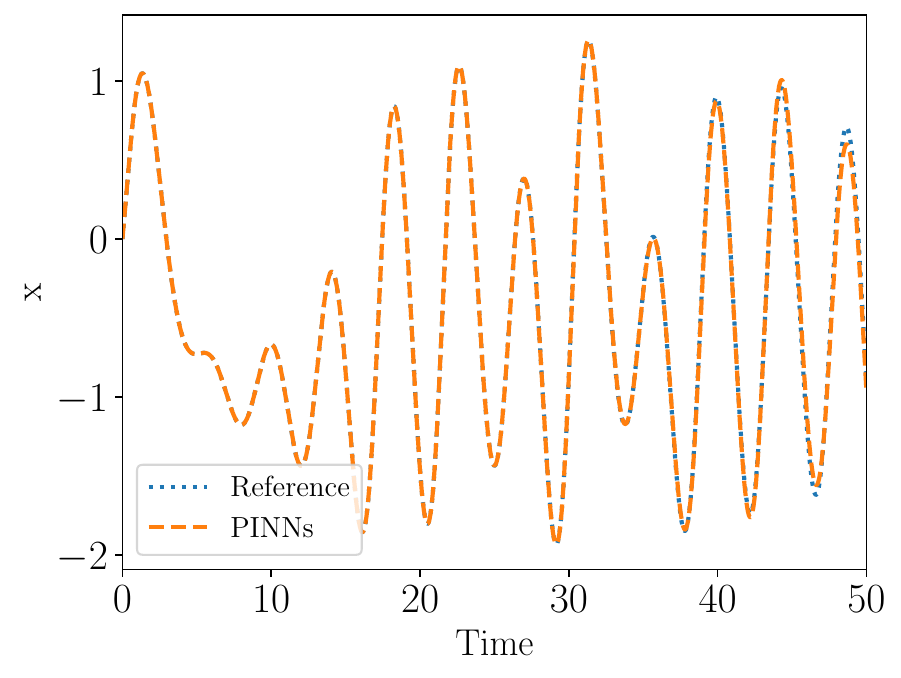} 
\end{tabular}
\caption{Comparison with HCS-PINN and SCS-PINN solutions of Jerky dynamics equation with chebfun solution \cite{driscoll2014chebfun}.}
\label{fig:jerk_eqn} 
\end{figure*}

\begin{figure} [H]
	\centering
	\includegraphics[width=0.6\textwidth]{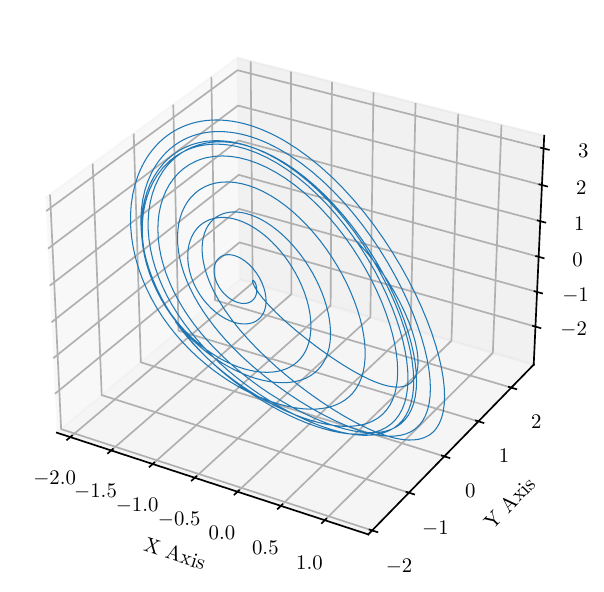}
	\caption{Evolution of Jerky dynamics in 3D phase space with HCSPINN ($nt=10$).}
	\label{fig:jerk_3D}
\end{figure}

The third order derivative in time requires that $C^2$ continuity should be enforced for the Jerk equation. A penalty parameter of $\lambda_I=10$ was used for SCS-PINN to improve convergence. Both HCS-PINN and SCS-PINN methods fail to converge for $nt=1$. The HCS-PINN iterations diverge and the line search method fails for L-BFGS optimizer resulting in a very high value of $L^2$ error for $nt=1$. The SCS-PINN iterations converges to a constant (or zero) solution even for $nt=5$, as shown in Figure \ref{fig:jerk_eqn}. Using the HCS-PINN method, we were able to obtain O($10^{-3}$) accurate results for $nt=5$ and $nt=10$ up to $t=50$. Figure \ref{fig:jerk_3D} shows the solution in a 3D phase space for $nt=10$ , which depicts the chaotic nature of its evolution. 

\begin{table}[h] 
\caption{Relative $L^2$ Error for Jerk equation (10,000 Adam iterations + 300 L-BFGS iterations, loss tolerance = $1 \times 10^{-7}$)}
\begin{center}
\begin{tabular}{ |c|c|c|c|c| } 
 \hline
 $nt$ & \thead{HCS-PINN} & \thead{HCS-PINN \\Time (secs)} & \thead{SCS-PINN} & \thead{SCS-PINN \\Time (secs)}\\  \hline
 1 & 8.4944 $\times$ 10$^{1} $& 299 & 1.7552 $\times$ 10$^{0}$& 1584 \\ 
 5 & 4.3303 $\times$ 10$^{-3}$& 1979 & 1.7543 $\times$ 10$^{0}$ & 1494\\ 
 10 & 1.0087 $\times$ 10$^{-3}$& 1791 & 2.8439 $\times$ 10$^{-2}$ & 1257 \\ 
 \hline
\end{tabular}
\end{center}
\end{table}

While some hyper-parameter tuning of the penalty parameter of the initial condition ($\lambda_I$) was conducted to ensure a fair comparison between HCS-PINN and SCS-PINN for all problems, it is feasible to manually adjust weights to achieve similar or better accuracy for SCS-PINN. We note that there are approaches to adaptively adjust weights for optimally balancing the loss \cite{wang2022and, xiang2022self}. However, this comes with additional computational expense. In this aspect, HCS-PINN is advantageous as it obviates the need to manually or automatically adjust the weights of time continuity constraints. 

\section{Summary}\label{sec:summary}
PINNs often struggle for time-dependent problems due to issues with temporal causality. 
This has previously been addressed by performing a time-domain decomposition, employing a distinct neural network for each time segment, and ensuring temporal continuity through terms incorporated into the loss function.  In this work we have introduced the HCS-PINN method, which eliminates the need to include these terms by strictly enforcing temporal continuity between successive neural networks.  

Both methods were demonstrated on first, second, and third order problems in time. The advection and wave equations are linear, hyperbolic problems for which PINNs often erroneously converge to the zero solution if the specified characteristic speeds are large. HCS-PINN has shown to be a superior approach in avoiding this spurious solution. Allen-Cahn and Korteweg-de Vries are both nonlinear problems which are widely used as benchmarks in the literature. In all cases, the hard constrained PINNs outperformed the soft constrained counterparts in terms of accuracy. 
Finally, a jerky dynamics problem was presented that is chaotic in nature, making it very sensitive to initial conditions. HCS-PINN was able to obtain the correct solution with only 5 time steps while SCS-PINN was not, and for 10 time steps its accuracy was better by an order of magnitude.

\section*{Acknowledgement} \label{acknow}
This work was performed under the auspices of the U.S. Department of Energy by Lawrence Livermore
National Laboratory under the Contract DE-AC52-07NA27344. 

\appendix
\section{Periodic Boundary Condition Enforcement} \label{Appendix:PBC}
Periodic boundary conditions are enforced via a method introduced by Dong and Ni~\cite{dong2021method}. This architecture involves passing the spatial input through a layer consisting of $C^m$ functions. For instance, for $C^{\infty}$ continuity, the following layer can be used:
\begin{equation}
    \text{Layer} = \{ sin(\omega x), cos(\omega x), ..., sin(k \omega x), cos(k \omega x) \}
\end{equation}
While using higher order terms can help capture higher modes \cite{wang2023expert, penwarden2023unified}, only the first two terms are used throughout this work, i.e. $k=1$. 

\section{Empirical Causal Weighting} \label{EmpWeight}
The temporal causality in enforced in this work by using multiple time segments and solving the problem sequentially. However, this does not preclude the use of other schemes to enforce causality within a time step. To this end, an empirical weighting scheme is used in which the loss function is multiplied by term that linearly decreases in time. In particular, the following form proposed by NVIDIA Modulus \cite{NvidiaModulus} is used:
\begin{equation}
    \lambda_P = C_T \left(1-\frac{t}{t_{\text{max}}} \right)+1
\end{equation}
With the exception of the Jerky dynamics problem, $t_{\text{max}}$ is taken to be the maximum time in the problem, not in the time step. For all problems, $C_T=10$ was used as causal weighting parameter. 
\section{Hyper-Parameters}
A list of hyper-parameters used in various problems is found in Table \ref{Tab:hyperparameters}.
\begin{table}[H]
\caption{Hyper-parameters for numerical experiments. NN denotes Neural network, MB denotes Mini-batch, and FB denotes Full-batch.}
\begin{center}
\begin{tabular}{ c|c|c|c|c|c|c|c} 
\Xhline{3\arrayrulewidth}
 Case & \thead{NN \\ depth} & \thead{NN \\ width} & \thead{Batch type\\ and size} & \thead{Adam \\step size} & \thead{Adam \\iteration}& \thead{L-BFGS \\iteration}& $\lambda_I$\\  \hline
 Advection & 4 & 32 & MB, 128 & 5$\times 10^{-3}$ & 10,000 & 1,000 & 1 \\ 
 Wave & 4 & 32 & MB, 128 & 5$\times 10^{-3}$ & 20,000 & 1,000 & 1 \\ 
 Allen-Cahn & 4 & 32 & MB, 256 & 5$\times 10^{-3}$ & 20,000 & 200 & 100  \\ 
 KdV & 4 & 32 & MB, 256 & 2$\times 10^{-3}$ & 20,000 & 200 & 100 \\
 Jerk &  3 & 16 & FB, 2001 & 2$\times 10^{-3}$ & 10,000 & 300 & 10  \\ 
\Xhline{3\arrayrulewidth}
\end{tabular}
\end{center} \label{Tab:hyperparameters}
\end{table}

\bibliographystyle{ieeetr}
\bibliography{PINNS_ref}

\begin{thebibliography}{10}

\bibitem{raissi}
M.~Raissi, P.~Perdikaris, and G.~E. Karniadakis, ``Physics-informed neural
  networks: A deep learning framework for solving forward and inverse problems
  involving nonlinear partial differential equations,'' {\em Journal of
  Computational physics}, vol.~378, pp.~686--707, 2019.

\bibitem{cai2021physics}
S.~Cai, Z.~Mao, Z.~Wang, M.~Yin, and G.~E. Karniadakis, ``Physics-informed
  neural networks ({PINNs}) for fluid mechanics: A review,'' {\em Acta
  Mechanica Sinica}, vol.~37, no.~12, pp.~1727--1738, 2021.

\bibitem{mahmoudabadbozchelou2022nn}
M.~Mahmoudabadbozchelou, G.~E. Karniadakis, and S.~Jamali, ``{nn-PINNs}:
  Non-newtonian physics-informed neural networks for complex fluid modeling,''
  {\em Soft Matter}, vol.~18, no.~1, pp.~172--185, 2022.

\bibitem{eivazi2022physics}
H.~Eivazi, M.~Tahani, P.~Schlatter, and R.~Vinuesa, ``Physics-informed neural
  networks for solving reynolds-averaged navier--stokes equations,'' {\em
  Physics of Fluids}, vol.~34, no.~7, 2022.

\bibitem{biswas2023three}
S.~K. Biswas and N.~Anand, ``Three-dimensional laminar flow using physics
  informed deep neural networks,'' {\em Physics of Fluids}, vol.~35, no.~12,
  2023.

\bibitem{cai2021heat}
S.~Cai, Z.~Wang, S.~Wang, P.~Perdikaris, and G.~E. Karniadakis,
  ``Physics-informed neural networks for heat transfer problems,'' {\em Journal
  of Heat Transfer}, vol.~143, no.~6, p.~060801, 2021.

\bibitem{ghaffari2023deep}
Y.~Ghaffari~Motlagh, P.~K. Jimack, and R.~de~Borst, ``Deep learning phase-field
  model for brittle fractures,'' {\em International Journal for Numerical
  Methods in Engineering}, vol.~124, no.~3, pp.~620--638, 2023.

\bibitem{sarma2023variational}
A.~Sarma, C.~Annavarapu, P.~Roy, S.~Jagannathan, and D.~Valiveti, ``Variational
  interface physics informed neural networks ({VI-PINNs}) for heterogeneous
  subsurface systems,'' in {\em ARMA US Rock Mechanics/Geomechanics Symposium},
  pp.~ARMA--2023, ARMA, 2023.

\bibitem{niaki2021physics}
S.~A. Niaki, E.~Haghighat, T.~Campbell, A.~Poursartip, and R.~Vaziri,
  ``Physics-informed neural network for modelling the thermochemical curing
  process of composite-tool systems during manufacture,'' {\em Computer Methods
  in Applied Mechanics and Engineering}, vol.~384, p.~113959, 2021.

\bibitem{tanyu2023deep}
D.~N. Tanyu, J.~Ning, T.~Freudenberg, N.~Heilenk{\"o}tter, A.~Rademacher,
  U.~Iben, and P.~Maass, ``Deep learning methods for partial differential
  equations and related parameter identification problems,'' {\em Inverse
  Problems}, vol.~39, no.~10, p.~103001, 2023.

\bibitem{yang2021b}
L.~Yang, X.~Meng, and G.~E. Karniadakis, ``{B-PINNs}: Bayesian physics-informed
  neural networks for forward and inverse {PDE} problems with noisy data,''
  {\em Journal of Computational Physics}, vol.~425, p.~109913, 2021.

\bibitem{lu2021physics}
L.~Lu, R.~Pestourie, W.~Yao, Z.~Wang, F.~Verdugo, and S.~G. Johnson,
  ``Physics-informed neural networks with hard constraints for inverse
  design,'' {\em SIAM Journal on Scientific Computing}, vol.~43, no.~6,
  pp.~B1105--B1132, 2021.

\bibitem{jagtap2022deep}
A.~D. Jagtap, D.~Mitsotakis, and G.~E. Karniadakis, ``Deep learning of inverse
  water waves problems using multi-fidelity data: Application to
  serre--green--naghdi equations,'' {\em Ocean Engineering}, vol.~248,
  p.~110775, 2022.

\bibitem{chen2020physics}
Y.~Chen, L.~Lu, G.~E. Karniadakis, and L.~Dal~Negro, ``Physics-informed neural
  networks for inverse problems in nano-optics and metamaterials,'' {\em Optics
  express}, vol.~28, no.~8, pp.~11618--11633, 2020.

\bibitem{serebrennikova2024physics}
A.~Serebrennikova, R.~Teubler, L.~Hoffellner, E.~Leitner, U.~Hirn, and
  K.~Zojer, ``Physics informed neural networks reveal valid models for reactive
  diffusion of volatiles through paper,'' {\em Chemical Engineering Science},
  vol.~285, p.~119636, 2024.

\bibitem{reddy2022finite}
J.~N. Reddy, N.~Anand, and P.~Roy, {\em Finite element and finite volume
  methods for heat transfer and fluid dynamics}.
\newblock Cambridge University Press, 2022.

\bibitem{krishnapriyan2021characterizing}
A.~Krishnapriyan, A.~Gholami, S.~Zhe, R.~Kirby, and M.~W. Mahoney,
  ``Characterizing possible failure modes in physics-informed neural
  networks,'' {\em Advances in Neural Information Processing Systems}, vol.~34,
  pp.~26548--26560, 2021.

\bibitem{wang2022respecting}
S.~Wang, S.~Sankaran, and P.~Perdikaris, ``Respecting causality is all you need
  for training physics-informed neural networks,'' {\em arXiv preprint
  arXiv:2203.07404}, 2022.

\bibitem{cuomo2022scientific}
S.~Cuomo, V.~S. Di~Cola, F.~Giampaolo, G.~Rozza, M.~Raissi, and F.~Piccialli,
  ``Scientific machine learning through physics--informed neural networks:
  Where we are and what’s next,'' {\em Journal of Scientific Computing},
  vol.~92, no.~3, p.~88, 2022.

\bibitem{dong2021method}
S.~Dong and N.~Ni, ``A method for representing periodic functions and enforcing
  exactly periodic boundary conditions with deep neural networks,'' {\em
  Journal of Computational Physics}, vol.~435, p.~110242, 2021.

\bibitem{sukumar2022exact}
N.~Sukumar and A.~Srivastava, ``Exact imposition of boundary conditions with
  distance functions in physics-informed deep neural networks,'' {\em Computer
  Methods in Applied Mechanics and Engineering}, vol.~389, p.~114333, 2022.

\bibitem{wight2020solving}
C.~L. Wight and J.~Zhao, ``Solving {Allen-Cahn} and {Cahn-Hilliard} equations
  using the adaptive physics informed neural networks,'' {\em arXiv preprint
  arXiv:2007.04542}, 2020.

\bibitem{mattey2022novel}
R.~Mattey and S.~Ghosh, ``A novel sequential method to train physics informed
  neural networks for {Allen Cahn} and {Cahn Hilliard} equations,'' {\em
  Computer Methods in Applied Mechanics and Engineering}, vol.~390, p.~114474,
  2022.

\bibitem{bihlo2022physics}
A.~Bihlo and R.~O. Popovych, ``Physics-informed neural networks for the
  shallow-water equations on the sphere,'' {\em Journal of Computational
  Physics}, vol.~456, p.~111024, 2022.

\bibitem{wang2023expert}
S.~Wang, S.~Sankaran, H.~Wang, and P.~Perdikaris, ``An expert's guide to
  training physics-informed neural networks,'' {\em arXiv preprint
  arXiv:2308.08468}, 2023.

\bibitem{penwarden2023unified}
M.~Penwarden, A.~D. Jagtap, S.~Zhe, G.~E. Karniadakis, and R.~M. Kirby, ``A
  unified scalable framework for causal sweeping strategies for
  physics-informed neural networks ({PINNs}) and their temporal
  decompositions,'' {\em arXiv preprint arXiv:2302.14227}, 2023.

\bibitem{jax2018github}
J.~Bradbury, R.~Frostig, P.~Hawkins, M.~J. Johnson, C.~Leary, D.~Maclaurin,
  G.~Necula, A.~Paszke, J.~Vander{P}las, S.~Wanderman-{M}ilne, and Q.~Zhang,
  {\em {JAX}: composable transformations of {P}ython+{N}um{P}y programs}, 2018.

\bibitem{kingma2014adam}
D.~P. Kingma and J.~Ba, ``Adam: A method for stochastic optimization,'' {\em
  arXiv preprint arXiv:1412.6980}, 2014.

\bibitem{liu1989limited}
D.~C. Liu and J.~Nocedal, ``On the limited memory {BFGS} method for large scale
  optimization,'' {\em Mathematical programming}, vol.~45, no.~1-3,
  pp.~503--528, 1989.

\bibitem{glorot2010understanding}
X.~Glorot and Y.~Bengio, ``Understanding the difficulty of training deep
  feedforward neural networks,'' in {\em Proceedings of the thirteenth
  international conference on artificial intelligence and statistics},
  pp.~249--256, JMLR Workshop and Conference Proceedings, 2010.

\bibitem{driscoll2014chebfun}
T.~A. Driscoll, N.~Hale, and L.~N. Trefethen, ``Chebfun guide,'' 2014.

\bibitem{braga2022characteristics}
U.~Braga-Neto, ``Characteristics-informed neural networks for forward and
  inverse hyperbolic problems,'' {\em arXiv preprint arXiv:2212.14012}, 2022.

\bibitem{allen1979microscopic}
S.~M. Allen and J.~W. Cahn, ``A microscopic theory for antiphase boundary
  motion and its application to antiphase domain coarsening,'' {\em Acta
  metallurgica}, vol.~27, no.~6, pp.~1085--1095, 1979.

\bibitem{miura1976korteweg}
R.~M. Miura, ``The {Korteweg--deVries} equation: a survey of results,'' {\em
  SIAM review}, vol.~18, no.~3, pp.~412--459, 1976.

\bibitem{lorenz1963deterministic}
E.~N. Lorenz, ``Deterministic nonperiodic flow,'' {\em Journal of atmospheric
  sciences}, vol.~20, no.~2, pp.~130--141, 1963.

\bibitem{smale1974differential}
S.~Smale and M.~W. Hirsch, {\em Differential equations, dynamical systems, and
  linear algebra}, vol.~60.
\newblock Elsevier, 1974.

\bibitem{rossler1979chaos}
O.~E. Rössler, ``Continuous chaos - four prototype equations,'' {\em Annals of
  the New York Academy of Sciences}, vol.~316, no.~1, pp.~376--392, 1979.

\bibitem{sprott1997some}
J.~Sprott, ``Some simple chaotic jerk functions,'' {\em American Journal of
  Physics}, vol.~65, no.~6, pp.~537--543, 1997.

\bibitem{eichhorn1998transformations}
R.~Eichhorn, S.~J. Linz, and P.~H{\"a}nggi, ``Transformations of nonlinear
  dynamical systems to jerky motion and its application to minimal chaotic
  flows,'' {\em Physical Review E}, vol.~58, no.~6, p.~7151, 1998.

\bibitem{sprott2009simplifications}
J.~Sprott, ``Simplifications of the {Lorenz} attractor,'' {\em Nonlinear
  dynamics, psychology, and life sciences}, vol.~13, no.~3, p.~271, 2009.

\bibitem{wang2022and}
S.~Wang, X.~Yu, and P.~Perdikaris, ``When and why {PINNs} fail to train: A
  neural tangent kernel perspective,'' {\em Journal of Computational Physics},
  vol.~449, p.~110768, 2022.

\bibitem{xiang2022self}
Z.~Xiang, W.~Peng, X.~Liu, and W.~Yao, ``Self-adaptive loss balanced
  physics-informed neural networks,'' {\em Neurocomputing}, vol.~496,
  pp.~11--34, 2022.

\bibitem{NvidiaModulus}
NVIDIA, Santa Clara, CA, {\em Nvidia Modulus}, 2022.

\end{thebibliography}

\end{document}